\documentclass[lettersize,journal]{IEEEtran}
\usepackage{amsmath,amsfonts}
\usepackage{algorithmic}
\usepackage{algorithm}
\usepackage{array}
\usepackage[caption=false,font=normalsize,labelfont=sf,textfont=sf]{subfig}
\usepackage{textcomp}
\usepackage{stfloats}
\usepackage{url}
\usepackage{verbatim}
\usepackage{graphicx}
\usepackage{cite}
\usepackage{multirow}
\usepackage{graphicx}
\usepackage{booktabs}
\usepackage{enumitem}
\usepackage{colortbl}
\usepackage{adjustbox}
\usepackage[commandnameprefix=always]{changes}

\usepackage{multirow}
\usepackage[table,xcdraw]{xcolor}
\usepackage[colorlinks, citecolor=green, linkcolor=red, citecolor=blue, urlcolor=cyan]{hyperref}
\usepackage[capitalise]{cleveref}
\begin{document}

\title{UMCFuse: A Unified Multiple Complex Scenes Infrared and Visible Image Fusion Framework}

\author{Xilai Li, Xiaosong Li, Tianshu Tan, Huafeng Li, Tao Ye
\thanks{This research was supported by the National Natural Science Foundation of China (No. 62201149),the Basic and Applied Basic Research of Guangdong Province (No. 2023A1515140077), the Natural Science Foundation of Guangdong Province (No.2024A1515011880), the National Natural Science Foundation of China (No. 52374166), the Research Fund of Guangdong-HongKong-Macao Joint Laboratory for Intelligent Micro-Nano Optoelectronic Technology (No. 2020B1212030010), and the Yunnan Fundamental Research Projects (Nos. 202301AV070004, 202501AS070123).
(Corresponding author: Xiaosong Li)}
\thanks{Xilai Li and Xiaosong Li are with the Guangdong-HongKong-Macao Joint Laboratory for Intelligent Micro-Nano Optoelectronic Technology, School of Physics and Optoelectronic Engineering, Foshan University, Foshan 528225, China. (emails: 20210300236@stu.fosu.edu.cn; lixiaosong@buaa.edu.cn).}
\thanks{Tianshu Tan is with the School of Engineering, Hong Kong University of Science and Technology, Hong Kong, China. (e-mail: ttanad@connect.ust.hk).}
\thanks{Huafeng Li is with the School of Information Engineering and Automation, Kunming University of Science and Technology, Kunming 650500, China. (e-mail: hfchina99@163.com).}
\thanks{Tao Ye is with the School of Mechanical Electronic and Information Engineering, China University of Mining and Technology, Beijing 100083, China. (e-mail: ayetao198715@163.com).}

}


\markboth{Journal of \LaTeX\ Class Files,~Vol.~14, No.~8, August~2021}%
{Shell \MakeLowercase{\textit{et al.}}: A Sample Article Using IEEEtran.cls for IEEE Journals}


\maketitle

\begin{abstract}
Infrared and visible image fusion has emerged as a prominent research area in computer vision. However, little attention has been paid to the fusion task in complex scenes, leading to sub-optimal results under interference. To fill this gap, we propose a unified framework for infrared and visible images fusion in complex scenes, termed UMCFuse. Specifically, we classify the pixels of visible images from the degree of scattering of light transmission, allowing us to separate fine details from overall intensity. Maintaining a balance between interference removal and detail preservation is essential for the generalization capacity of the proposed method. Therefore, we propose an adaptive denoising strategy for the fusion of detail layers. Meanwhile, we fuse the energy features from different modalities by analyzing them from multiple directions. Extensive fusion experiments on real and synthetic complex scenes datasets cover adverse weather conditions, noise, blur, overexposure, fire, as well as downstream tasks including semantic segmentation, object detection, salient object detection, and depth estimation, consistently indicate the superiority of the proposed method compared with the recent representative methods. Our code is available at \href{https://github.com/ixilai/UMCFuse}{https://github.com/ixilai/UMCFuse}.

\end{abstract}


\section{Introduction}
\label{sec:intro}
Infrared and visible image fusion (IVIF) techniques \cite{r1,r2,r3,r4,r5,r77,r78,R86,r102,r103} are utilized to enhance the quality of captured images by amalgamating complementary information from both spectral ranges. In well-lit conditions, visible imaging captures fine details and rich color information. However, visible imaging is sensitive to environmental interference in low-light environments, leading to detail loss and increased noise. In contrast, infrared imaging, while lacking color and fine details, excels at detecting temperature variations and revealing hidden objects through thermal radiation. Additionally, it reliably captures thermal information, even in low-light or nighttime conditions. Thus, each modality offers unique advantages, and their fusion compensates for individual limitations while leveraging complementary strengths across diverse conditions. Fused images improve performance in downstream tasks such as semantic segmentation \cite{r6,r7}, object detection \cite{r9,r10}, salient object detection \cite{r11,r12}, and depth estimation \cite{r13,r14}.

Existing IVIF methods can be broadly categorized into deep learning (DL)-based \cite{r15,r16,r17,r19,r20,r21} and traditional methods \cite{r22,r23,r24,r25,r26,r27,r28}. DL-based methods are roughly divided into auto-encoder (AE)-based \cite{r2,r29,r30}, generative adversarial network (GAN)-based \cite{r3,r31,r32,r33}, and convolutional neural network (CNN)-based methods \cite{r34,r35,r36,r80,r79}. Specifically, AE-based methods employ encoders and decoders for feature extraction and image reconstruction. Encoders extract features from infrared or visible images, which are then processed by shared decoders for deep feature learning. In contrast, GAN-based methods model the image fusion task as an adversarial process: the generator combines features from two source images, while the discriminator guides the enhancement of informative details. Meanwhile, CNN-based methods automatically extract features from input images using hierarchical structures. 
On the other hand, traditional methods encompass multi-scale transform (MST)-based methods \cite{r83,R84}, sparse representation (SR)-based  methods \cite{r75,r82}, and others \cite{r76}. MST-based methods typically decompose the image into a series of subbands, extract crucial pixel information at various scale spaces, and subsequently derive the fusion results through inverse transformation. By contrast, SR differs from traditional MST by learning from a collection of training images for image fusion, thereby gaining a more stable and meaningful representation of the source images. 

\subsection{Motivation}
Although recent methods \cite{r2,r3,r4,r17,r22,r37} have shown excellent performance, a critical limitation warrants attention. Most existing IVIF techniques implicitly assume that image fusion is performed in high-quality scenes, such as normal weather and uniform lighting. However, real-world applications, including autonomous driving, scene analysis, and environmental monitoring, frequently encounter complex environmental interferences like extreme weather, motion blur, noise, overexposure, and fire. Such interferences may occur either individually or in combination, presenting significant challenges for robust image fusion. One possible solution to address such challenges is to incorporate image restoration techniques as pre-processing steps for fusion algorithms. However, this approach is hindered by the lack of comprehensive image restoration models capable of achieving unified weights across diverse scenarios, making it difficult to accurately detect and address specific complex scenes. Moreover, employing a separate restoration model for each disturbance significantly amplifies computational complexity.

\begin{figure}[t]
  \centering
   \includegraphics[width=1.0\linewidth]{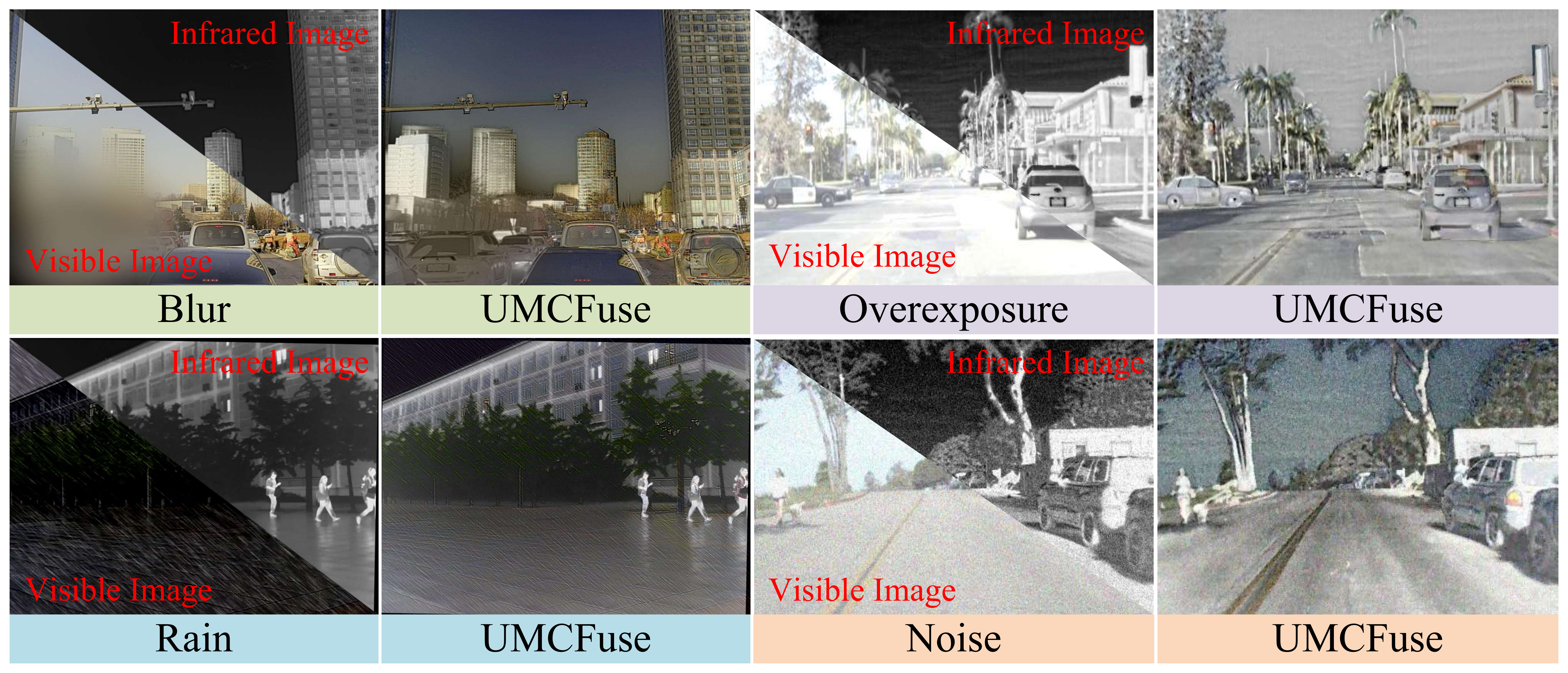}
   \caption{Fusion results of the proposed algorithm in multiple complex scenes, where Blur represents a scene where the sensor is destroyed by raindrops.}
   \label{fig1}
\end{figure}

While numerous algorithms have been designed to address specific adverse conditions, such as haze \cite{r64}, noise \cite{r65,r66}, and overexposure \cite{r67}, they exhibit two key challenges: (1) Their frameworks are often tailored for a specific type of interference in complex scenes, thus restrict their generalization ability to diverse and dynamic real-world scenarios. (2) They may prioritize the removal of interfering pixels, overlooking multi-modality feature extraction, leading to suboptimal fusion performance, especially in standard scenarios.
In summary, the above methods \cite{r64,r65,r66,r67} exhibit substantial limitations when deployed in practical applications.

\cref{fig1} shows the fusion results of several complex scenes. These practical challenges necessitate the development of IVIF technology capable of simultaneously mitigating multiple environmental interferences while achieving high-quality multi-modality image fusion. Motivated by these challenges, we propose an IVIF method for a complex environment with multiple interferences. Compared with existing methods, the proposed framework has two key advantages. First, it transcends the ``restoration + fusion'' paradigm by unifying interference removal and feature extraction within a single framework, improving computational efficiency and adaptability. Second, unlike existing fusion methods targeting a single type of disturbance (e.g., haze or overexposure), the proposed method is explicitly designed to handle multiple simultaneous disturbances (e.g., haze, noise, overexposure, fire), enhancing its practical applicability in real-world scenarios. Specifically, we design a transmission map-based decomposition model to decouple various degradations in complex scenes. The resulting structure and contrast layers are further decomposed into high and low frequencies using a novel edge-preserving filter. On this basis, we construct a fusion strategy of high and low frequencies from the perspectives of adaptive denoising and multi-directional feature detection, respectively.
\textbf{To the best of our knowledge, this is the first unified framework explicitly designed to address IVIF challenges in complex and dynamic scenes.}
Our main contributions as follows:

\begin{enumerate}[label=\arabic*),left=0pt]
\item We propose a novel framework designed to address challenging scenes. The proposed method adaptively identifies and processes interfering information in the source images, facilitating high-quality fusion even in extreme environments such as haze, noise, heavy rain, and fire.

\item A light transmission-aware decomposition model is proposed that can accurately classify interfering pixels within complex scenes. Moreover, we develop an adaptive denoising scheme for high-frequency images, achieving a balance between interference removal and detail preservation.

\item We validate the effectiveness of the proposed algorithm under real and synthetic data. Moreover, we further analyze the feasibility of the proposed algorithm in a wider range of real complex scenes.
\end{enumerate}

\begin{figure*}[t]
  \centering
   \includegraphics[width=1.0\linewidth]{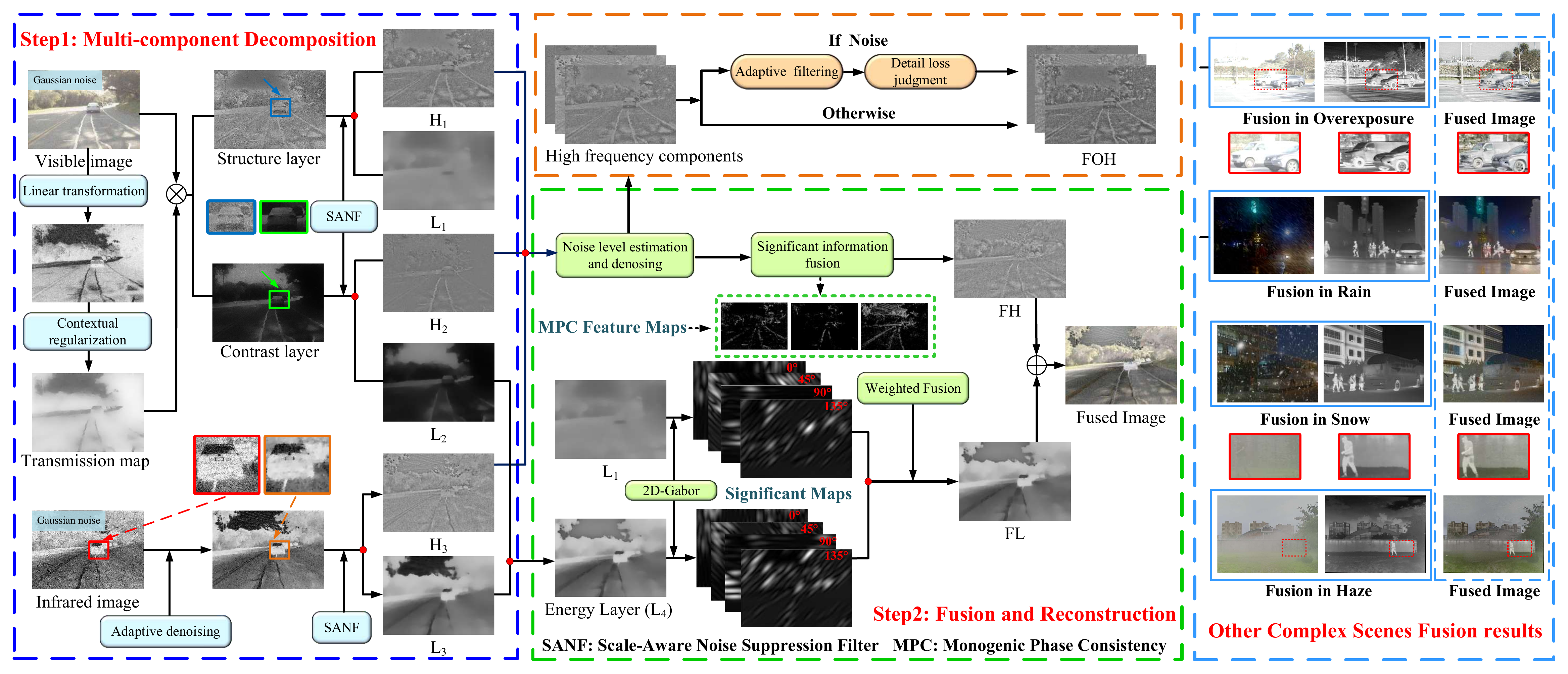}
   \caption{Overall framework of the proposed method, which contains two main steps: multi-component decomposition, feature fusion, and reconstruction. The rightmost region shows four complex scenes fusion examples.}
   \label{fig2}
\end{figure*}
\section{Related Work}
\subsection{Existing Infrared and Visible Image Fusion Methods}
In recent years, DL has advanced the field of IVIF, with many researchers focusing on DL-based methods. Some DL-based algorithms \cite{r2,r16,r34,r37,r38,r39,r98} have achieved promising performance in capturing multi-modality image information, and they usually design complex neural network structures and loss functions according to the needs of a specific task. For example, Zhao et al. \cite{r2} proposed a correlation-based two-branch feature decomposition network by combining CNN and Transformer. Cheng et al. \cite{r99} supervised the training of multi-modality image fusion algorithm using low-level vision tasks from digital photographic fusion, eliminating the need to rely on high-level semantic information. However, such data-driven algorithms may fail when confronted with unseen data and often lack strong generalization ability.

Owing to its high degree of interpretability, the traditional methods \cite{r22} have achieved impressive achievements in the field of image processing over many years. MST, a significant branch of traditional methods\cite{r72,r73}, can consider the distribution of feature information by utilizing the multi-scale and directional information of an image.
With the gradual improvement of a single IVIF task, many IVIF algorithms \cite{r4,r17,r81,R85,r41,r42,r43,r44,r71,r95,r97} challenge partial downstream tasks (semantic segmentation, object detection, etc.) and complex scenes (haze, noise, and overexposure, etc.) by giving the algorithms multi-tasking capabilities. This trend is expected to shape the future development of IVIF. For example, to address the inconsistency between the optimization objectives of image fusion and downstream tasks, Wu et al. \cite{r100} leveraged the semantic information from the segment anything model to guide the training of the fusion network. Furthermore, to enhance the performance of fusion results on downstream tasks, Bai et al. \cite{r101} proposed a learnable fusion loss that allows the fusion network to be supervised by downstream task losses in a meta-learning framework.
However, due to the lack of data and the increased training difficulty, almost no algorithm is able to handle multiple extreme scenes simultaneously in adverse conditions.
Owing to the rigorous mathematical theory support, the traditional IVIF model has the potential to partially fill this gap, as it does not require extensive training data. In this context, we can uniformly regard pixels that hinder the distribution of details in complex scenes and affect human visual perception as interference pixels. Nonetheless, current research in this area is insufficient.

\section{Proposed complex scenes fusion model}
This section describes the flow of the proposed algorithm in detail. The overall process is illustrated in \cref{fig2}.
\subsection{Multi-component Decomposition}
Traditional image fusion methods often rely on high- and low-frequency decompositions, employing techniques such as edge-preserving filtering \cite{r75} and multi-scale transforms \cite{r83}. In contrast, DL-based methods \cite{r2,r90} utilize specialized feature extraction modules to identify shared and unique features for decomposition. However, both techniques depend solely on the internal priors of the image, assuming that an image can be effectively represented as a combination of high-frequency and low-frequency components. Moreover, they do not leverage external prior knowledge, such as estimating optical transmission maps using the Atmospheric Scattering Model (ASM) for image decomposition.

\subsubsection{Transmission Map Estimation}
The ASM provides a comprehensive framework for elucidating light propagation phenomena across different atmospheric conditions \cite{r68,r69}. While initially applied predominantly in dehazing tasks, the ASM can also applies to other challenging scenarios, including overexposure, snow, rain, and fire scenes. For instance, overexposure results from excessive light concentration, while snowflakes introduce reflection and refraction complexities. Similarly, fire scenes involve light scattering and absorption by smoke and particulates, resulting in a hazy, orange-tinted atmosphere. Given that these scenes involve common atmospheric phenomena (scattering, absorption, and transmission), we assert that the ASM can provide a unified description for these conditions.

\noindent\textbf{\textit{Step 1. Coarse transmission map acquisition}}

\noindent Take dehazing task for example, assuming equal transmittance and atmospheric light across color channels in the input visible source image, the smallest channel can more accurately identify the noise level \cite{r45}. The transmission map $t$ is then calculated by:
\begin{equation}
  t(x,y) = \frac{{{minA} - {minI}(x, y)}}{{{minA} - {minJ}(x, y)}}
  \label{eq1}
\end{equation}
\noindent Here, ${minI}(x, y)$, ${minJ}(x, y)$, and ${minA}$ represent the minimum intensity values for the hazy image, the clear image, and the atmospheric light, respectively. A minimum filter is applied to implement this process. The conventional ASM focuses on defogging by recovering a clear image $J$ from a foggy image $I$. However, in real-world scenarios, challenging conditions such as overexposure, snow, and rain, also affect light scattering, posing greater challenges in $J$ recovery. Rather than recovering $J$, the proposed algorithm focuses solely on decomposing images based on light transmittance. To reduce the impact of $J$ estimation errors on the transmittance map $t$, we reformulated the ASM as:
\begin{equation}
t(x, y) = \left(\min A - \min I(x,y)\right)^{\frac{\min A}{\min A - \min J(x,y)}}
  \label{eq3}
\end{equation}
where ${minA}$ is solved by dark channel prior \cite{r62}. If $minJ(x, y)$ is known, $t(x,y)$ can be computed. Due to the positive correlation between ${minI}(x, y)$ and $minJ(x, y)$ \cite{r46}, $minJ(x, y)$ can be written as: 
 \begin{equation}
{minJ}(x, y) = \frac{1}{{\ln({minA} - {minI}(x, y))^\beta}}
  \label{eq4}
\end{equation}
To find ${minJ}(x, y)$, we determine $\beta$ using the least squares fitting, which minimizes the sum of squared errors. We train the proposed algorithm with the NH-Haze dataset \cite{r47} of 50 image pairs, haze and haze-free. According to the least squares algorithm, we calculate the value of $\beta$ as $1.2778$.

\noindent\textbf{\textit{Step 2. Transmission map optimization}}

\noindent The transmission map, obtained from scene pixel depth estimation, is sensitive to errors and noise. To improve its robustness, we employ a contextual regularization-based image smoothing function \cite{r48} to refine the transmission map:
 \begin{equation}
\frac{\lambda}{2} \left\|t - \hat{t}\right\|^2_2 + \sum_{j \in \omega} \left\|W_j \circ (D_j \otimes t)\right\|_1
  \label{eq5}
\end{equation}
where $\hat{t}$ represents the smoothed transmission map, $\lambda$ stands for the regularization parameter, $\omega$ is an index set \cite{r48}, $D_j$ denotes the differential operator, $\circ$ represents the element-wise multiplication, $\otimes$ signifies the convolution operator, and $W_j$ is the weight matrix of the filter \cite{r48}. To improve the smoothing function, we adaptively adjust $\lambda$ based on noise level measurement \cite{r49}. The noise level $\sigma$ is estimated based on non-local self-similarity prior \cite{r49}. $\lambda = 1.5 \times e^{-\sigma}$. This adaptation decreases computational complexity and retains edge details while effectively suppressing noise.

\begin{figure}[t]
  \centering
   \includegraphics[width=0.95\linewidth]{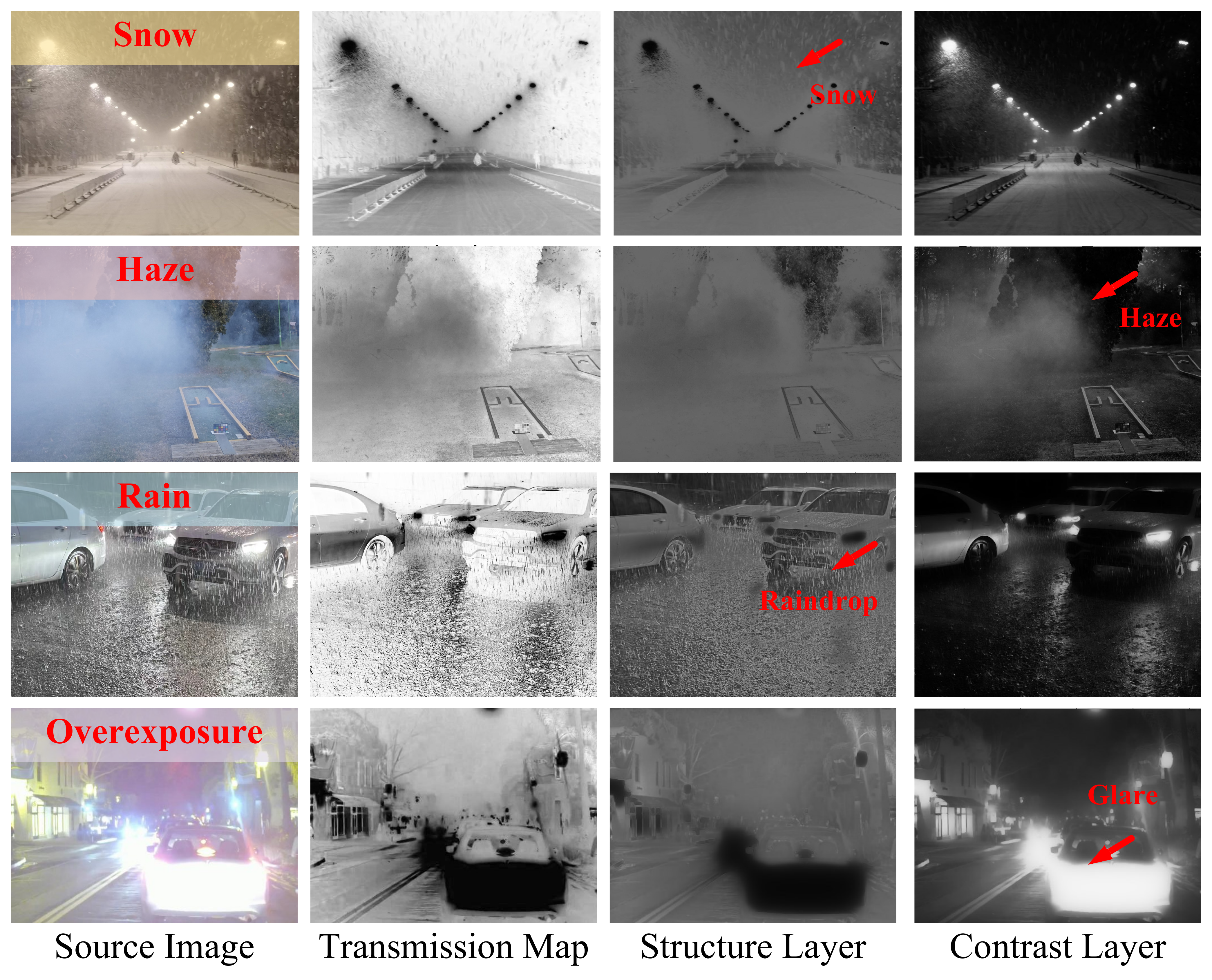}
   \caption{Visualization for effectiveness analysis of decomposition strategy.}
   \label{fig3}
\end{figure}
\subsubsection{Contrast and Structure Layers Decomposition}

Transmission maps capture the propagation of light through a scene, accounting for effects such as scattering and refraction. As shown in \cref{fig3}, the transmission map is used to identify and partition areas of light attenuation in the source image, enabling the separate handling of degraded pixels and original content. For example, during rainy day imaging, raindrops scatter and refract light, resulting in pixel value variations within the transmission map. The refracted light reaching the camera no longer originates from a specific point in the scene but from a broader area encompassing the raindrop. As a result, the transmission map captures the blurring effect caused by light passing through the raindrop. Similarly, in snow scenes, light scattering and refraction from snowflakes blur the transmission map and reduce image contrast. This phenomenon allows snow features to be captured in the transmission map and separated into the structural layer. In overexposed scenes, excessive light intensity leads to a loss of detail in bright regions, producing large uniform areas in the transmission map with pixel values near saturation. 

Separating degradation components reduces the interference they introduce during feature extraction. When too much detailed information in the image overlaps with the degradation factors, they may have similar gradients or pixel values, which can cause the feature detection operator to be unable to distinguish the useful information from the interfering information. Thus, separating degradation using the transmission map significantly reduces interference during feature extraction.
We use the transmission map to decompose the visible image $I_{\text{vis}}$ into the contrast layer ($CL$) and structure layer ($SL$).
 \begin{equation}
CL(x, y) = I_{\text{vis}}(x, y) \times (1 - t(x, y))
  \label{eq6}
\end{equation}
 \begin{equation}
SL(x, y) = I_{\text{vis}}(x, y) \times t(x, y)
  \label{eq7}
\end{equation}
This strategy distinguishes elements in a complex scene (e.g., snow, haze, raindrops), aiding in the subsequent suppression and processing of interfering pixels.

\subsubsection{High and Low Frequency Decomposition}
Edge-preserving filters suppress noise while retaining edges. For example, Zhou et al. \cite{r50} achieved this via a weighted least-squares framework. However, noise in the source image can mask feature information at higher scales, impacting scale perception metrics. To solve this, we use the Sobel operator for gradient detection in eight directions and generate new weights. This strategy optimizes the basic framework of the above and we define the new filter as a Scale-Aware Noise Suppression Filter (SANF). SANF is a scale-aware suppression filter designed to adaptively remove noise at different scales. It effectively preserves the edge information of objects in the image, striking a balance between maintaining image details (e.g., textures and contours) and structure (e.g., object edges and shapes). SANF can be represented as follows:
\begin{equation}
\begin{aligned}
\small
& \sum_{p} \left( (O^d (p) - I(p))^2 \right. \\
& + \frac{\kappa}{\xi_p^{d-1} G_p^{d-1} Q_{r,p} S_p^{d-1} + \epsilon} \left( (\partial_x O^d)_p^2 + (\partial_y O^d)_p^2 \right) \biggr)
\end{aligned}
\label{eq77}
\end{equation}
where $p$ represents each pixel, $O^d$ signifies the filtered result after the $d$-th iteration, with $d$ taking the value of $3$, $Q_{r,p}$ is the scale-aware metric used to suppress image structures with spatial scales smaller than $r$, $G_p^{d-1}$ is the guidance gradient, $\xi_p^{d-1}$ is the weight employed to penalize the guidance gradient $G_p^{d-1}$, thereby suppressing instances where the filter output gradient at pixel $p$ exceeds that of the input image. $\epsilon=0.0001$. More detailed information about this filter can be found in \cite{r50}. Additionally, $S_p$ denotes the noise suppression weight, which is calculated as follows.
\begin{equation}
S_p = \sqrt{\sum_{D=1}^8 \frac{1}{8} \times GL_{p}^D}
  \label{eq9}
\end{equation}
where $D$ denotes the number of Sobel operator templates in various directions, and $GL$ is the gradient map detected by the Sobel operator across different directions. Eq.~\eqref{eq77} defines a weighted least squares objective, iteratively optimized by updating smoothing weights based on gradient differences. The output image $O$ represents the low-frequency component, and the high-frequency component is the residual between the input and the output. Consequently, the SANF process can be represented by $\text{SANF}(I, \kappa, r)$, where $r$ is set to $1$, and the weight $\kappa$ is used to modulate the degree of image smoothing.

Considering their lower resolution and susceptibility to noise from weak signals and atmospheric interference, infrared images are filtered with SANF using a modified parameter to emphasize noise reduction over texture preservation and enhance fusion quality. Assuming that the noise level $\sigma_{ir}$ of the infrared image is greater than $0$, we filter it using the SANF, and $\kappa = 0.01 \times \log_5(\sigma_{ir})$.
After obtaining $CL$, $SL$, and the pre-processed infrared image $\bar{I}_{ir}$, we decompose them using SANF into low-frequency ($L_\alpha$) and high-frequency ($H_\alpha$) components, where $\alpha \in \{1, 2, 3\}$ corresponds to $SL$, $CL$ and $\bar{I}_{ir}$, respectively, and $\kappa = \frac{\kappa'}{\max\left(e^{0.03 \sigma_{\alpha}}\right)}$, and $\kappa'=0.4$.

\subsection{Fusion and Reconstruction}
\subsubsection{High-Frequency Fusion}
In adverse conditions, images suffer from various types of interference. High-frequency components, which are responsible for details, are particularly sensitive to environmental changes and noise. Thus, high-frequency fusion involves three steps:

\noindent\textbf{\textit{Step 1. Adaptive Denoising}}

\noindent Conventional filtering methods often rely on manually setting parameters to accommodate different noise levels, which both increases the complexity of use and may affect performance. Filtering that is too strong leads to loss of detail, while too weak does not allow for effective denoising. For this reason, we adaptively adjust the filter strength based on the estimated noise level in the image, and dynamically control the parameter $\delta_{\alpha}$ that determines the output gradient to achieve a balance between noise suppression and detail preservation.

First, we calculate the noise level $\sigma_{\alpha}^{h}$ of the high-frequency component, based on noise levels, we decide whether to filter interfering pixels, preserving essential details and edges.
\begin{equation}
OH_{\alpha} = \begin{cases}
  \text{SANF}(H_{\alpha}, \delta_{\alpha}, r), & \text{if } \sigma_{\alpha}^h > 0 \\
  H_{\alpha}, & \text{otherwise}
\end{cases}
  \label{eq10}
\end{equation}
where $OH_{\alpha}$ denotes the optimized high-frequency component, $\delta_{\alpha} = 0.4 \times \log_5 (\sigma_{\alpha}^h)$, if $\sigma_{\alpha}^h = 0$, $H_{\alpha}$ is directly used as the final optimized high-frequency component $FOH_{\alpha}$.

\noindent\textbf{\textit{Step 2. Detail Loss Judgment}}

\noindent While image filters are effective at reducing noise, they often lead to a loss of detail in high-frequency components. Additionally, manually designed rules fall short in assessing the degree of detail loss and its impact on visual quality post-denoising. To assess detail loss, we use Kullback–Leibler (KL) divergence \cite{r51} to measure the difference between denoising and original images in terms of detail. A smaller KL divergence means better preservation of details and edge information in denoised high-frequency components.

Let $\rho_{\alpha}^1$ and $\rho_{\alpha}^0$ denote two probability distributions of $H_\alpha$ and $OH_\alpha$, respectively, the KL divergence is calculated as follows:
 \begin{equation}
KL(\rho_{\alpha}^1 \| \rho_{\alpha}^0) = \int \rho_{\alpha}^1(R) \log \left(\frac{\rho_{\alpha}^1(R)}{\rho_{\alpha}^0(R)}\right) \, dR
  \label{eq11}
\end{equation}
where $R$ represents the random variable values. By assessing the magnitude of KL divergence, we assess the similarity of grayscale detail distributions between the images. Based on this, $FOH_{\alpha}$ is calculated as follows:
 \begin{equation}
FOH_{\alpha} = \begin{cases}
  H_{\alpha}, & \text{if } KL > 0.05 \\
  OH_{\alpha}, & \text{otherwise}
\end{cases}
  \label{eq12}
\end{equation}

\begin{figure}[t]
  \centering
   \includegraphics[width=1\linewidth]{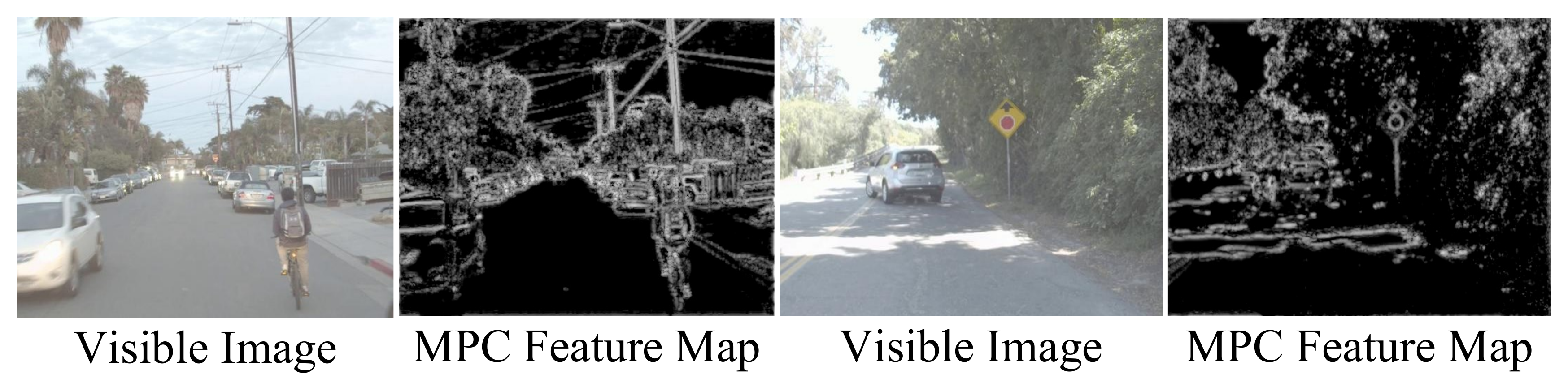}
   \caption{MPC feature extraction visualization results.}
   \label{figmpc}
\end{figure}

\noindent\textbf{\textit{Step 3. Significant Information Fusion}}

\noindent The Monogenic phase consistency (MPC) \cite{r52} is a superior feature extraction operator that provides precise local phase information, in contrast to traditional phase consistency methods, which are susceptible to noise. The results of the MPC feature extraction on visible images are demonstrated in \cref{figmpc}. The MPC is formulated as follows:
\begin{equation}
\begin{aligned}
\small
M H_{\alpha}(x, y) = & \omega_{\alpha}(x, y) \left\lfloor 1 - \gamma \times a \cos\left(\frac{E_{\alpha}(x, y)}{A_{\alpha}(x, y)}\right) \right\rfloor \\
& \times \left\lfloor E_{\alpha}(x, y) - T \right\rfloor / (A_{\alpha}(x, y) + 1e-4)
\end{aligned}
\end{equation}
where $E_{\alpha}(x,y)$ and $A_{\alpha}(x,y)$ are local energy and amplitude sums of $FOH_\alpha(x,y)$, $\omega_{\alpha}(x,y)$ and $T$ represent the weight function and noise compensation factor respectively, and $\gamma$ is set to $1.5$. Next, we introduce weighted fusion rules for obtaining the final high-frequency components $FH$:
 \begin{equation}
FH(x,y)=\sum_{\alpha=1}^{3}\left(\frac{MH_{\alpha}(x,y)+\eta}{\max\left(\sum_{\alpha=1}^{3}MH_{\alpha}(x,y)\right)+\eta}\times H_{\alpha}\right)
  \label{eq14}
\end{equation}
where $\eta=0.1$ is to prevent the denominator from being $0$.
\subsubsection{Low-Frequency Fusion}
First, we extract energy information from the low-frequency components of the contrast layer and the infrared image, obtaining the energy layer $L_4$:
 \begin{equation}
L_4 = L_2 \times \text{Map}_E + L_3 \times (1 - \text{Map}_E)
  \label{eq15}
\end{equation}
where
\begin{equation}
\text{Map}_E = \begin{cases}
  1, & \text{if } L_2 > L_3 \\
  0, & \text{otherwise}
\end{cases}
  \label{eq16}
\end{equation}

To capture low-frequency feature information, we propose a new fusion rule based on the Gabor filter \cite{r53}. The Gabor filter is a versatile linear filter combining Gaussian and sinusoidal functions for feature extraction across scales and directions. The 2D-Gabor function is defined as:
\begin{equation}
g_{\theta} = \exp\left(-\frac{{x^{\prime 2} + \tau^2 y^{\prime 2}}}{{2 \hat{\sigma}^2}}\right) \cos\left(\frac{{2\pi x^{\prime}}}{{\hat{\lambda}}} + \psi\right)
  \label{eq17}
\end{equation}
where $ x' = x \cos\theta + y \sin\theta $ and $ y^{\prime} = -x \sin\theta + y \cos\theta $, $\hat{\lambda}=20$ denotes the wavelength of the sine, $\theta$ denotes the direction of the parallel strips in the Gabor filter kernel, $\theta \in \{0^\circ,45^\circ,90^\circ,135^\circ\}$, $\tau=0.5$ denotes the spatial aspect ratio, $\psi=0$ denotes the phase shift, and $\hat{\sigma}$ is the standard deviation of the Gaussian envelope.
\begin{figure*}[t]
  \centering
   \includegraphics[width=1\linewidth]{fig5.pdf}
   \caption{Qualitative comparison of all methods on normal, noise and overexposure scenes, respectively.}
   \label{figs1}
\end{figure*}

Subsequently, we can extract the pixel information in four different directions with the Gabor filter:
\begin{equation}
h_{\gamma}^{\theta}(x, y) = L_{\gamma}(x, y) * g_{\theta}
  \label{eq18}
\end{equation}
where $\gamma \in \{1,4\}$, $*$ is the convolution operator, and $h_{\gamma}^{\theta}$ is the filtering results in different directions. Then, the significant information can be extracted by calculating the variance $V_{\gamma}^{\theta}$ in all directions in $h_{\gamma}^{\theta}$ to construct the weight map:
\begin{equation}
V_{\theta}^{\gamma}(x, y) = \frac{1}{m \times n} \sum_{x=1}^m \sum_{y=1}^n (h_{\theta}^{\gamma}(x, y) - \bar{h}_{\theta}^{\gamma}(x, y))^2
  \label{eq19}
\end{equation}
where $m$ and $n$ denote the image length and width, respectively, and $\bar{h}_{\theta}^{\gamma}$ denotes the mean value of $h_{\theta}^{\gamma}$. The fusion weight map $W_{\gamma}^L$ can be calculated as:
\begin{equation}
W_{\gamma}^L = \frac{V_{\gamma}^L e^{EN(L_{\gamma})}}{e^{EN(L_1)} \times V_1^L + e^{EN(L_4)} \times V_4^L}
  \label{eq20}
\end{equation}
where $V_{\gamma}^L$ represents the average value of $V_{\theta}^{\gamma}$ in all directions and $EN(\cdot)$ represents the image entropy operator \cite{r54}. Finally, the low-frequency component fusion result $FL$ can be obtained according to the fusion weight map $W_{\gamma}^L$:
\begin{equation}
FL = W_1^L \times L_1 + W_4^L \times L_4
  \label{eq21}
\end{equation}
\subsubsection{Fusion Result Reconstruction}
According to $FH$ and $FL$, the final fusion result $F$ can be obtained: 
\begin{equation}
F = FH + FL
  \label{eq22}
\end{equation}

\begin{figure*}[h]
  \centering
   \includegraphics[width=1\linewidth]{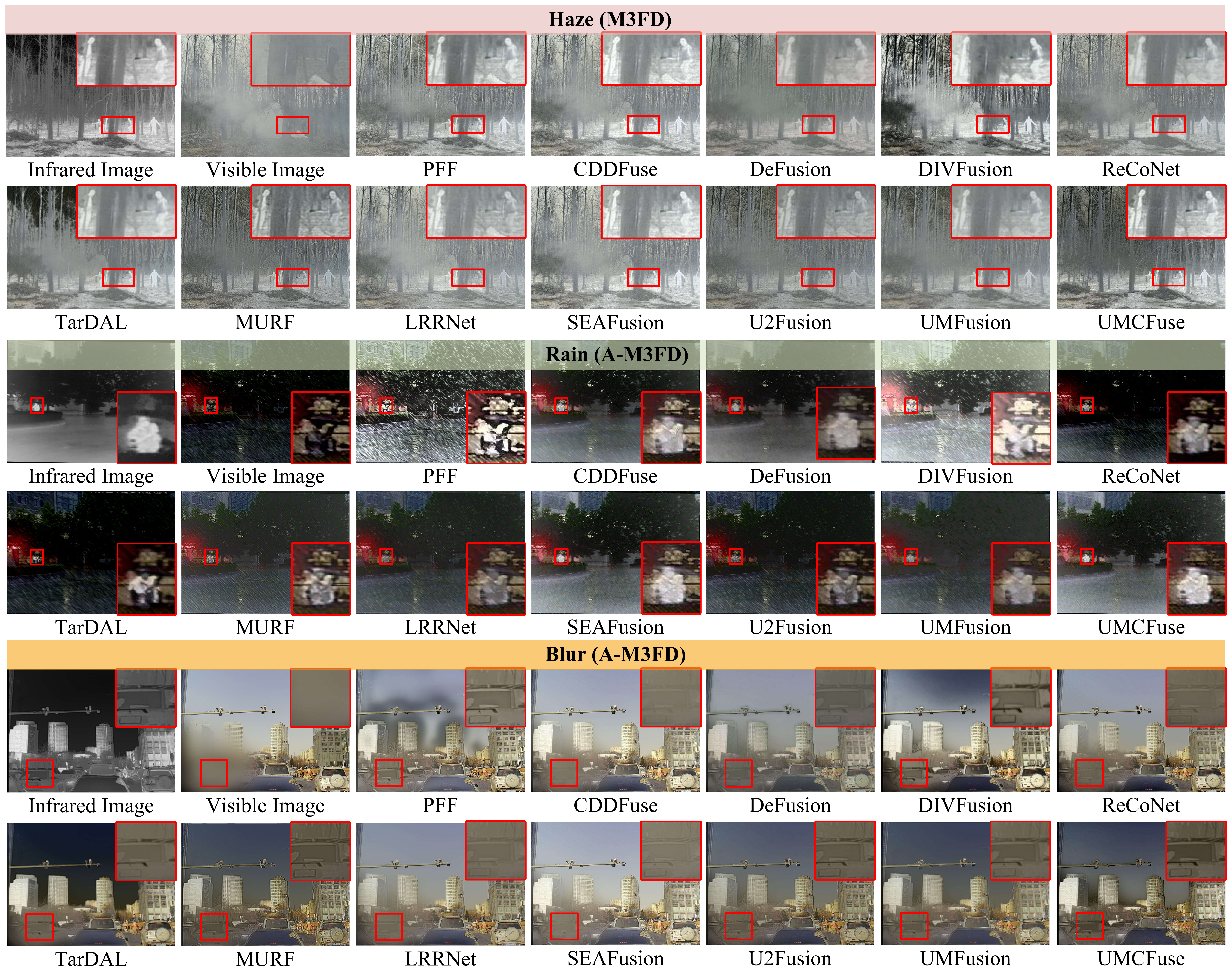}
   \caption{Qualitative comparison of all methods on three complex scenes: haze, rain and blur.}
   \label{figs2}
\end{figure*}

\begin{figure*}[h]
  \centering
   \includegraphics[width=1\linewidth]{fig7.pdf}
   \caption{Qualitative comparison of all methods on three complex scenes: snow, noise and rain, and fire.}
   \label{figs3}
\end{figure*}

\section{Experiment}
\subsection{Experimental setup}
We used the RoadScene \cite{r34} and M3FD \cite{r3} datasets, which include challenging real-world scenes such as smoke, strong light, and rain. Additionally, we selected image pairs depicting fire scenes from the TNO dataset \cite{r55}. To train the DL-based methods, we used 1,200 image pairs from three widely used datasets: 170 pairs from RoadScene, 1,000 pairs from M3FD, and 30 pairs from TNO. The corresponding test set includes 50 pairs from RoadScene, 300 from M3FD, and 20 from TNO. Furthermore, we evaluated performance using eight popular metrics: $Q_{MI}$, $Q_{NCIE}$, $Q_{G}$, $Q_{abf}$, $Q_{CV}$ \cite{r56}, $VIF$ \cite{r87}, $EN$, and $SSIM$. With the exception of $Q_{CV}$, higher values indicate better fusion performance. For comparison, we selected 11 state-of-the-art methods, namely PFF \cite{r22}, CDDFuse \cite{r2}, DeFusion \cite{r39}, DIVFusion \cite{r17}, ReCoNet \cite{r41}, TarDAL \cite{r3}, MURF \cite{r42}, LRRNet \cite{r16}, SEAFusion \cite{r4}, U2Fusion \cite{r34}, and UMFusion \cite{r20}.

\subsection{Qualitative comparison}
We add Gaussian noise (standard deviation of 20) on both visible and infrared images from the RoadScene dataset, creating A-RoadScene. B-RoadScene simulates conditions with strong interference from bright light sources. We constructed A-M3FD by adding rain and snow \cite{r57} to visible images based on M3FD. Additionally, B-M3FD replicates challenging scenes with Gaussian noise (standard deviation of 100) in rainy scene. It is noteworthy that the M3FD dataset inherently includes real rain and haze within its scenes, whereas the RoadScene dataset comprises genuine overexposed scenes. Our degradation of the dataset is intended to enhance the interference effect and increase the difficulty of the test and the amount of data.

\noindent \textbf{Noise Scene Fusion:} In noisy scenes (\cref{figs1}), fusion results vary significantly across algorithms: PFF, MURF, LRRNet, U2Fusion, and UMFusion struggle with contrast and fail to preserve key infrared luminance. PFF and DIVFusion amplify source image noise, as seen in the red zoomed-in region, obscuring fine structural details such as vehicle edges. CDDFuse and SEAFusion compensate for weak infrared textures by assigning higher weights to the visible image, but this introduces significant noise from the visible image into the fusion process. 
In contrast, the proposed algorithm demonstrates superior performance in preserving infrared image luminance, visible image texture, and reducing noisy pixels.

\noindent \textbf{Overexposure Scene Fusion:} In the B-RoadScene dataset presented in \cref{figs1} , we evaluate various algorithms under overexposure scene. In B-RoadScene, overexposure affects PFF, CDDFuse, ReCoNet, and SEAFusion, resulting in the loss of vehicle target information. DIVFusion, MURF, LRRNet, and U2Fusion all properly mitigate the effects of overexposed pixels to make targets easier to recognize in the scene. They fill in a large number of infrared pixels in the fusion result, adjusting the brightness of the whole image. However, their images can deviate from natural visual perception, similar to the sky area are changed to gray and color distortion. The proposed method effectively addresses the challenge of overexposure by balancing the energy of visible images and the thermal radiation of infrared images. This results in the preservation of detailed information and a reasonable contrast.

\noindent \textbf{Haze Scene Fusion:} In \cref{figs2}, we assess the performance of different algorithms in scenes affected by haze, where the visibility in visible images is severely reduced. In the magnified red area, all methods successfully incorporate target information from infrared images. However, ReCoNet, MURF, U2Fusion, and UMFusion exhibit fusion results with poor contrast. PFF, DIVFusion, and SEAFusion fail to accurately identify haze as interfering information. The presence of haze pixel information during the fusion process further affects the distribution of weak texture information within the infrared images.

\noindent \textbf{Rain Scene Fusion:} In complex rainy scenes (see \cref{figs2,figs3}), existing algorithms still face significant challenges. In the zoomed red area, PFF, DIVFusion, MURF, LRRNet, and U2Fusion encounter difficulties in capturing target information, struggling to differentiate interfering pixels in the visible image. Combining rain-line noise with Gaussian noise presents an even more significant challenge in B-M3FD. With the exception of DeFusion and UMCFuse, other methods struggle to maintain a reasonable level of contrast across the scene. In \cref{figs2}, we show the scene with water-damaged visible sensors (labeled as Blur). Most methods suffer from severe blur, except PFF, DIVFusion, MURF, and the proposed method.

\noindent \textbf{Snow Scene Fusion:} The snow scene in \cref{figs3} displays fusion results from various algorithms. Snow particles are uniformly distributed in the visible image scene, which results in some target information being obscured and interfered. In the highlighted red areas, the DIVFusion algorithm amplifies the snow effects in the scene, indicating its inability to effectively distinguish regions with valuable details. In contrast, while the DeFusion, MURF, U2Fusion, and UMFusion methods suppress snow interference, they also tend to diminish overall scene contrast. This suggests that when encountering unknown interference, these algorithms may opt to discard target information rather than selectively removing the interference and restoring the original details. The proposed algorithm achieves superior results by effectively mitigating snow interference while preserving target details and scene contrast.

\noindent \textbf{Fire Scene Fusion:} In the fire scene comparison (\cref{figs3}), the objective is to accurately locate the fire source and maintain reasonable image contrast. PFF, CDDFuse, and SEAFusion keep excessive visible image information, resulting in overall high brightness and difficulty in fire detection. In contrast, the proposed algorithm effectively balances both modalities.

\begin{table*}[t]
\caption{Quantitative comparison of all methods on four metrics $Q_{MI}$, $Q_{NCIE}$, $Q_{G}$ and $Q_{abf}$, with ``Average RoadScene" and ``Average M3FD" denoting the average scores on three datasets each. Green is the best, blue is the second, and red is the third.}
\centering
\begin{adjustbox}{width=\textwidth}
\begin{tabular}{c|cccc|cccc|cccc}
\hline
                          & \multicolumn{4}{c|}{\textbf{Average RoadScene}}                                                                                                                                                                                                                                              & \multicolumn{4}{c|}{\textbf{Average M3FD}}                                                                                                                                                                                                                                                   & \multicolumn{4}{c}{\textbf{TNO}}                                                                                                                                                                                                                                                             \\ \cline{2-13} 
\multirow{-2}{*}{Methods} & $Q_{MI}$↑                                                                        & $Q_{NCIE}$↑                                                                      & $Q_{G}$↑                                                                         & $Q_{abf}$↑                                                  & $Q_{MI}$↑                                                                        & $Q_{NCIE}$                                                                      & $Q_{G}$↑                                                                         & $Q_{abf}$↑                                                  & $Q_{MI}$↑                                                                        & $Q_{NCIE}$↑                                                                      & $Q_{G}$↑                                                                         & $Q_{abf}$↑                                                  \\ \hline
PFF \cite{r22}                       & \multicolumn{1}{c|}{0.2704}                                                & \multicolumn{1}{c|}{0.8049}                                                & \multicolumn{1}{c|}{0.2976}                                                & 0.3565                                                & \multicolumn{1}{c|}{0.1691}                                                & \multicolumn{1}{c|}{0.8038}                                                & \multicolumn{1}{c|}{0.1981}                                                & 0.3031                                                & \multicolumn{1}{c|}{0.4183}                                                & \multicolumn{1}{c|}{0.8052}                                                & \multicolumn{1}{c|}{0.1795}                                                & 0.3680                                                \\
CDDFuse \cite{r2}                   & \multicolumn{1}{c|}{0.3335}                                                & \multicolumn{1}{c|}{0.8058}                                                & \multicolumn{1}{c|}{0.3017}                                                & 0.3563                                                & \multicolumn{1}{c|}{\cellcolor[HTML]{FFCCC9}{\color[HTML]{000000} 0.3230}} & \multicolumn{1}{c|}{\cellcolor[HTML]{D9E1F2}{\color[HTML]{000000} 0.8069}} & \multicolumn{1}{c|}{\cellcolor[HTML]{FFCCC9}{\color[HTML]{000000} 0.2770}} & 0.3349                                                & \multicolumn{1}{c|}{\cellcolor[HTML]{D9E1F2}{\color[HTML]{000000} 0.7746}} & \multicolumn{1}{c|}{\cellcolor[HTML]{FFCCC9}{\color[HTML]{000000} 0.8133}} & \multicolumn{1}{c|}{\cellcolor[HTML]{9AFF99}\textbf{0.3312}}               & 0.4301                                                \\
DeFusion \cite{r39}                  & \multicolumn{1}{c|}{0.3576}                                                & \multicolumn{1}{c|}{\cellcolor[HTML]{FFCCC9}{\color[HTML]{000000} 0.8062}} & \multicolumn{1}{c|}{0.2707}                                                & 0.3177                                                & \multicolumn{1}{c|}{\cellcolor[HTML]{9AFF99}\textbf{0.5266}}               & \multicolumn{1}{c|}{0.8050}                                                & \multicolumn{1}{c|}{\cellcolor[HTML]{D9E1F2}{\color[HTML]{000000} 0.2780}} & \cellcolor[HTML]{FFCCC9}{\color[HTML]{000000} 0.3414} & \multicolumn{1}{c|}{\cellcolor[HTML]{FFCCC9}{\color[HTML]{000000} 0.7344}} & \multicolumn{1}{c|}{0.8116}                                                & \multicolumn{1}{c|}{0.2612}                                                & \cellcolor[HTML]{FFCCC9}{\color[HTML]{000000} 0.4306} \\
DIVFusion \cite{r17}                 & \multicolumn{1}{c|}{0.3489}                                                & \multicolumn{1}{c|}{0.8060}                                                & \multicolumn{1}{c|}{0.2501}                                                & 0.2947                                                & \multicolumn{1}{c|}{0.2750}                                                & \multicolumn{1}{c|}{0.8049}                                                & \multicolumn{1}{c|}{0.2054}                                                & 0.3079                                                & \multicolumn{1}{c|}{0.6136}                                                & \multicolumn{1}{c|}{0.8087}                                                & \multicolumn{1}{c|}{0.2348}                                                & 0.3638                                                \\
ReCoNet \cite{r41}                   & \multicolumn{1}{c|}{\cellcolor[HTML]{9AFF99}\textbf{0.4097}}               & \multicolumn{1}{c|}{0.8067}                                                & \multicolumn{1}{c|}{0.2514}                                                & 0.3040                                                & \multicolumn{1}{c|}{0.2882}                                                & \multicolumn{1}{c|}{0.8050}                                                & \multicolumn{1}{c|}{0.2173}                                                & 0.3225                                                & \multicolumn{1}{c|}{0.7185}                                                & \multicolumn{1}{c|}{0.8105}                                                & \multicolumn{1}{c|}{0.2703}                                                & 0.3886                                                \\
TarDAL \cite{r3}                    & \multicolumn{1}{c|}{\cellcolor[HTML]{D9E1F2}{\color[HTML]{000000} 0.3957}} & \multicolumn{1}{c|}{\cellcolor[HTML]{D9E1F2}{\color[HTML]{000000} 0.8070}} & \multicolumn{1}{c|}{0.2844}                                                & 0.3295                                                & \multicolumn{1}{c|}{\cellcolor[HTML]{D9E1F2}{\color[HTML]{000000} 0.3244}} & \multicolumn{1}{c|}{0.8059}                                                & \multicolumn{1}{c|}{0.1880}                                                & 0.2235                                                & \multicolumn{1}{c|}{0.6161}                                                & \multicolumn{1}{c|}{0.8069}                                                & \multicolumn{1}{c|}{0.1455}                                                & 0.3192                                                \\
MURF \cite{r42}                     & \multicolumn{1}{c|}{0.3135}                                                & \multicolumn{1}{c|}{0.8057}                                                & \multicolumn{1}{c|}{0.3417}                                                & 0.3784                                                & \multicolumn{1}{c|}{0.2128}                                                & \multicolumn{1}{c|}{0.8042}                                                & \multicolumn{1}{c|}{0.1720}                                                & 0.2356                                                & \multicolumn{1}{c|}{0.6615}                                                & \multicolumn{1}{c|}{0.8091}                                                & \multicolumn{1}{c|}{0.3067}                                                & 0.4199                                                \\
LRRNet \cite{r16}                    & \multicolumn{1}{c|}{0.3525}                                                & \multicolumn{1}{c|}{0.8060}                                                & \multicolumn{1}{c|}{0.2588}                                                & 0.2986                                                & \multicolumn{1}{c|}{0.2649}                                                & \multicolumn{1}{c|}{0.8049}                                                & \multicolumn{1}{c|}{0.2256}                                                & 0.3287                                                & \multicolumn{1}{c|}{0.6630}                                                & \multicolumn{1}{c|}{0.8084}                                                & \multicolumn{1}{c|}{0.2275}                                                & 0.3436                                                \\
SEAFusion \cite{r4}               & \multicolumn{1}{c|}{0.3374}                                                & \multicolumn{1}{c|}{0.8059}                                                & \multicolumn{1}{c|}{0.3126}                                                & 0.3705                                                & \multicolumn{1}{c|}{0.3011}                                                & \multicolumn{1}{c|}{\cellcolor[HTML]{FFCCC9}{\color[HTML]{000000} 0.8064}} & \multicolumn{1}{c|}{0.2662}                                                & \cellcolor[HTML]{D9E1F2}{\color[HTML]{000000} 0.3442} & \multicolumn{1}{c|}{\cellcolor[HTML]{9AFF99}\textbf{0.7937}}               & \multicolumn{1}{c|}{\cellcolor[HTML]{D9E1F2}{\color[HTML]{000000} 0.8136}} & \multicolumn{1}{c|}{\cellcolor[HTML]{FFCCC9}{\color[HTML]{000000} 0.3153}} & \cellcolor[HTML]{D9E1F2}{\color[HTML]{000000} 0.4423} \\
U2Fusion  \cite{r34}                & \multicolumn{1}{c|}{0.3259}                                                & \multicolumn{1}{c|}{0.8057}                                                & \multicolumn{1}{c|}{\cellcolor[HTML]{D9E1F2}{\color[HTML]{000000} 0.3705}} & \cellcolor[HTML]{D9E1F2}{\color[HTML]{000000} 0.3955} & \multicolumn{1}{c|}{0.2407}                                                & \multicolumn{1}{c|}{0.8047}                                                & \multicolumn{1}{c|}{0.2548}                                                & 0.3243                                                & \multicolumn{1}{c|}{0.6484}                                                & \multicolumn{1}{c|}{0.8078}                                                & \multicolumn{1}{c|}{0.2318}                                                & 0.3969                                                \\
UMFusion \cite{r20}                 & \multicolumn{1}{c|}{0.3632}                                                & \multicolumn{1}{c|}{0.8067}                                                & \multicolumn{1}{c|}{\cellcolor[HTML]{FFCCC9}{\color[HTML]{000000} 0.3574}} & \cellcolor[HTML]{FFCCC9}{\color[HTML]{000000} 0.3827} & \multicolumn{1}{c|}{0.2695}                                                & \multicolumn{1}{c|}{0.8053}                                                & \multicolumn{1}{c|}{0.2273}                                                & 0.2784                                                & \multicolumn{1}{c|}{0.7170}                                                & \multicolumn{1}{c|}{0.8100}                                                & \multicolumn{1}{c|}{0.2019}                                                & 0.3741                                                \\ \hline
UMCFuse                   & \multicolumn{1}{c|}{\cellcolor[HTML]{FFCCC9}{\color[HTML]{000000} 0.3703}} & \multicolumn{1}{c|}{\cellcolor[HTML]{9AFF99}\textbf{0.8072}}               & \multicolumn{1}{c|}{\cellcolor[HTML]{9AFF99}\textbf{0.3991}}               & \cellcolor[HTML]{9AFF99}\textbf{0.4172}               & \multicolumn{1}{c|}{0.3166}                                                & \multicolumn{1}{c|}{\cellcolor[HTML]{9AFF99}\textbf{0.8070}}               & \multicolumn{1}{c|}{\cellcolor[HTML]{9AFF99}\textbf{0.2793}}               & \cellcolor[HTML]{9AFF99}\textbf{0.3457}               & \multicolumn{1}{c|}{0.6989}                                                & \multicolumn{1}{c|}{\cellcolor[HTML]{9AFF99}\textbf{0.8418}}               & \multicolumn{1}{c|}{\cellcolor[HTML]{D9E1F2}{\color[HTML]{000000} 0.3277}} & \cellcolor[HTML]{9AFF99}\textbf{0.4459}               \\ \hline
\end{tabular}
\label{tab1}
\end{adjustbox}
\end{table*}
\begin{table*}[h]
\caption{Quantitative comparison of all methods on four metrics $Q_{CV}$, $VIF$, $EN$, and $SSIM$, with ``Average RoadScene" and ``Average M3FD" denoting the average scores on three datasets each. Green is the best, blue is the second, and red is the third.}
\centering
\begin{adjustbox}{width=\textwidth}
\begin{tabular}{c|cccc|cccc|cccc}
\hline
                                   & \multicolumn{4}{c|}{\textbf{Average RoadScene}}                                                                                                                                                                               & \multicolumn{4}{c|}{\textbf{Average M3FD}}                                                                                                                                                                                    & \multicolumn{4}{c}{\textbf{TNO}}                                                                                                                                                                                              \\ \cline{2-13} 
\multirow{-2}{*}{\textbf{Methods}} & $Q_{CV}$↓                                                   & $VIF$↑                                                   & $EN$↑                                                     & $SSIM$↑                                                  & $Q_{CV}$↓                                                   & $VIF$↑                                                   & $EN$↑                                                    & $SSIM$↑                                                  &  $Q_{CV}$↓                                                  & $VIF$↑                                                   & $EN$↑                                                    & $SSIM$↑                                                  \\ \hline
PFF \cite{r22}                               & \multicolumn{1}{c|}{1690.4}                                                & \multicolumn{1}{c|}{0.2024}                                                & \multicolumn{1}{c|}{\cellcolor[HTML]{FFCCC9}{\color[HTML]{000000} 7.4566}}                                                & 0.3510                                                & \multicolumn{1}{c|}{1044.8}                                                & \multicolumn{1}{c|}{0.0970}                                                & \multicolumn{1}{c|}{\cellcolor[HTML]{FFCCC9}{\color[HTML]{000000} 7.3611}} & 0.1577                                                & \multicolumn{1}{c|}{914.17}                                                & \multicolumn{1}{c|}{0.1316}                                                & \multicolumn{1}{c|}{\cellcolor[HTML]{FFCCC9}{\color[HTML]{000000} 6.8567}} & 0.1699                                                \\
CDDFuse \cite{r2}                            & \multicolumn{1}{c|}{\cellcolor[HTML]{D9E1F2}{\color[HTML]{000000} 761.25}} & \multicolumn{1}{c|}{0.1700}                                                & \multicolumn{1}{c|}{6.8385}                                                & \cellcolor[HTML]{9AFF99}\textbf{0.5654}               & \multicolumn{1}{c|}{567.01}                                                & \multicolumn{1}{c|}{0.1291}                                                & \multicolumn{1}{c|}{7.3107}                                                & \cellcolor[HTML]{FFCCC9}{\color[HTML]{000000} 0.4866} & \multicolumn{1}{c|}{419.34}                                                & \multicolumn{1}{c|}{0.4046}                                                & \multicolumn{1}{c|}{6.5072}                                                & 0.3385                                                \\
DeFusion \cite{r39}                           & \multicolumn{1}{c|}{783.40}                                                & \multicolumn{1}{c|}{0.2827}                                                & \multicolumn{1}{c|}{6.6795}                                                & 0.3674                                                & \multicolumn{1}{c|}{\cellcolor[HTML]{D9E1F2}{\color[HTML]{000000} 555.91}} & \multicolumn{1}{c|}{\cellcolor[HTML]{D9E1F2}{\color[HTML]{000000} 0.1729}} & \multicolumn{1}{c|}{6.9442}                                                & 0.2552                                                & \multicolumn{1}{c|}{906.92}                                                & \multicolumn{1}{c|}{0.4125}                                                & \multicolumn{1}{c|}{6.2754}                                                & 0.2975                                                \\
DIVFusion \cite{r17}                          & \multicolumn{1}{c|}{1662.2}                                                & \multicolumn{1}{c|}{0.1552}                                                & \multicolumn{1}{c|}{\cellcolor[HTML]{D9E1F2}{\color[HTML]{000000} 7.5333}} & 0.2674                                                & \multicolumn{1}{c|}{867.14}                                                & \multicolumn{1}{c|}{0.1013}                                                & \multicolumn{1}{c|}{\cellcolor[HTML]{D9E1F2}{\color[HTML]{000000} 7.5887}} & 0.1937                                                & \multicolumn{1}{c|}{1361.2}                                                & \multicolumn{1}{c|}{0.1969}                                                & \multicolumn{1}{c|}{\cellcolor[HTML]{9AFF99}\textbf{7.0416}}               & 0.2797                                                \\

ReCoNet  \cite{r41}                         & \multicolumn{1}{c|}{869.21}                                                & \multicolumn{1}{c|}{\cellcolor[HTML]{D9E1F2}{\color[HTML]{000000} 0.3351}} & \multicolumn{1}{c|}{6.1088}                                                & 0.3107                                                & \multicolumn{1}{c|}{600.87}                                                & \multicolumn{1}{c|}{0.1176}                                                & \multicolumn{1}{c|}{6.6053}                                                & 0.1658                                                & \multicolumn{1}{c|}{\cellcolor[HTML]{9AFF99}\textbf{301.82}}               & \multicolumn{1}{c|}{0.3838}                                                & \multicolumn{1}{c|}{6.4273}                                                & 0.2542                                                \\
TarDAL \cite{r3}                             & \multicolumn{1}{c|}{1979.1}                                                & \multicolumn{1}{c|}{0.2377}                                                & \multicolumn{1}{c|}{6.9656}                                                & 0.3450                                                & \multicolumn{1}{c|}{1257.4}                                                & \multicolumn{1}{c|}{0.1542}                                                & \multicolumn{1}{c|}{6.5270}                                                & 0.1825                                                & \multicolumn{1}{c|}{467.88}                                                & \multicolumn{1}{c|}{0.3580}                                                & \multicolumn{1}{c|}{4.8048}                                                & 0.1638                                                \\
MURF \cite{r42}                             & \multicolumn{1}{c|}{1420.2}                                                & \multicolumn{1}{c|}{0.2286}                                                & \multicolumn{1}{c|}{7.0467}                                                & 0.4230                                                & \multicolumn{1}{c|}{765.57}                                                & \multicolumn{1}{c|}{0.0851}                                                & \multicolumn{1}{c|}{6.7986}                                                & 0.1687                                                & \multicolumn{1}{c|}{1148.4}                                                & \multicolumn{1}{c|}{\cellcolor[HTML]{D9E1F2}{\color[HTML]{000000} 0.4713}} & \multicolumn{1}{c|}{5.9744}                                                & \cellcolor[HTML]{D9E1F2}{\color[HTML]{000000} 0.3891} \\
LRRNet \cite{r16}                            & \multicolumn{1}{c|}{942.38}                                                & \multicolumn{1}{c|}{0.1925}                                                & \multicolumn{1}{c|}{7.0502}                                                & \cellcolor[HTML]{D9E1F2}{\color[HTML]{000000} 0.5023} & \multicolumn{1}{c|}{739.97}                                                & \multicolumn{1}{c|}{0.1452}                                                & \multicolumn{1}{c|}{6.7286}                                                & \cellcolor[HTML]{9AFF99}\textbf{0.5276}               & \multicolumn{1}{c|}{\cellcolor[HTML]{FFCCC9}{\color[HTML]{000000} 380.55}} & \multicolumn{1}{c|}{0.3362}                                                & \multicolumn{1}{c|}{5.8224}                                                & 0.1857                                                \\
SEAFusion \cite{r4}                         & \multicolumn{1}{c|}{\cellcolor[HTML]{FFCCC9}{\color[HTML]{000000} 773.34}} & \multicolumn{1}{c|}{0.1811}                                                & \multicolumn{1}{c|}{6.5998}                                                & 0.3380                                                & \multicolumn{1}{c|}{\cellcolor[HTML]{FFCCC9}{\color[HTML]{000000} 558.21}} & \multicolumn{1}{c|}{0.1378}                                                & \multicolumn{1}{c|}{7.3046}                                                & 0.2408                                                & \multicolumn{1}{c|}{548.48}                                                & \multicolumn{1}{c|}{0.3712}                                                & \multicolumn{1}{c|}{6.5171}                                                & \cellcolor[HTML]{FFCCC9}{\color[HTML]{000000} 0.3566} \\

U2Fusion \cite{r34}                          & \multicolumn{1}{c|}{1637.0}                                                & \multicolumn{1}{c|}{0.2904}                                                & \multicolumn{1}{c|}{6.6641}                                                & 0.4541                                                & \multicolumn{1}{c|}{675.21}                                                & \multicolumn{1}{c|}{0.1402}                                                & \multicolumn{1}{c|}{6.7348}                                                & 0.2094                                                & \multicolumn{1}{c|}{608.87}                                                & \multicolumn{1}{c|}{0.4411}                                                & \multicolumn{1}{c|}{5.5210}                                                & 0.2945                                                \\
UMFusion \cite{r20}                          & \multicolumn{1}{c|}{1549.2}                                                & \cellcolor[HTML]{FFCCC9}{\color[HTML]{000000}0.3051}                                                & \multicolumn{1}{c|}{6.8207}                                                & 0.4299                                                & \multicolumn{1}{c|}{790.00}                                                & \multicolumn{1}{c|}{\cellcolor[HTML]{9AFF99}\textbf{0.1754}}               & \multicolumn{1}{c|}{6.6427}                                                & 0.2389                                                & \multicolumn{1}{c|}{823.50}                                                & \multicolumn{1}{c|}{\cellcolor[HTML]{9AFF99}\textbf{0.5109}}               & \multicolumn{1}{c|}{5.9127}                                                & 0.2804                                                \\ \hline
UMCFuse                            & \multicolumn{1}{c|}{\cellcolor[HTML]{9AFF99}\textbf{747.17}}               & \multicolumn{1}{c|}{\cellcolor[HTML]{9AFF99}\textbf{0.3462}}               & \multicolumn{1}{c|}{\cellcolor[HTML]{9AFF99}{\color[HTML]{000000} \textbf{7.5407}}} & \cellcolor[HTML]{FFCCC9}{\color[HTML]{000000} 0.4662} & \multicolumn{1}{c|}{\cellcolor[HTML]{9AFF99}\textbf{543.61}}               & \cellcolor[HTML]{FFCCC9}{\color[HTML]{000000} 0.1584}                                                & \multicolumn{1}{c|}{\cellcolor[HTML]{9AFF99}\textbf{7.6365}}               & \cellcolor[HTML]{D9E1F2}{\color[HTML]{330001} 0.5194} & \multicolumn{1}{c|}{\cellcolor[HTML]{D9E1F2}{\color[HTML]{000000} 372.02}} & \multicolumn{1}{c|}{\cellcolor[HTML]{FFCCC9}{\color[HTML]{000000} 0.4645}} & \multicolumn{1}{c|}{\cellcolor[HTML]{D9E1F2}{\color[HTML]{000000} 6.9398}} & \cellcolor[HTML]{9AFF99}\textbf{0.3949}               \\ \hline
\end{tabular}
\label{tab2}
\end{adjustbox}
\end{table*}

\subsection{Quantitative Comparison}
\cref{tab1} and \cref{tab2} present the quantitative comparison results of all algorithms on seven datasets. The ``Average RoadScene'' represents the average scores across three datasets, RoadScene, A-RoadScene, and B-RoadScene. ``Average M3FD'' represents the average scores across three datasets, M3FD, A-M3FD, and B-M3FD. The TNO dataset specifically evaluates performance on fire scenes. 
The analysis of \cref{tab1} and \cref{tab2} reveals the following observations:

\begin{enumerate}[label=\arabic*),left=0pt]
\item Average RoadScene Dataset: The proposed algorithm demonstrates improvements of 7.72\%, 5.49\%, 1.85\%, and 3.31\% in the $Q_{G}$, $Q_{abf}$, $Q_{CV}$, and $VIF$ metrics, respectively, compared to the second-ranked method.

\item Average M3FD Dataset: The proposed algorithm shows improvements of 0.47\%, 0.44\%, 2.21\%, and 0.63\% in the $Q_{G}$, $Q_{abf}$, $Q_{CV}$, and $EN$ metrics, respectively.

\item TNO Dataset: The proposed algorithm exhibits improvements of 3.47\%, 3.93\%, 0.81\%, and 1.49\% in the $Q_{NCIE}$, $Q_{G}$, $Q_{abf}$, and $SSIM$ metrics, respectively.

\item Overall Comparison: When compared with the average performance of the 11 competing methods, the proposed algorithm achieves an overall improvement of 13.64\% in the $Q_{MI}$, $Q_{NCIE}$, $Q_{G}$, and $Q_{abf}$ metrics and an overall improvement of 34.06\% in the $Q_{CV}$, $VIF$, $EN$, and $SSIM$ metrics.
\end{enumerate}
\begin{table}[t]
\caption{Quantitative Comparison of All Methods on Object Detection and Semantic Segmentation Tasks.}
\begin{tabular}{c|cccccc}
\hline
\rowcolor{gray!20} Metrics                              & PFF   & CDD                           & DeF                           & DIV   & ReCo  & TarDAL                                 \\ \hline
{{ mAP@0.5}} & 0.804 & \cellcolor[HTML]{FFCCC9}0.816 & 0.798                         & 0.778 & 0.795 & 0.741                                  \\
mIoU                                 & 72.76 & 71.03                         & 74.78                         & 63.47 & 64.72 & \cellcolor[HTML]{FFCCC9}74.84          \\ \hline
\rowcolor{gray!20} Metrics                              & MURF  & LRR                           & SeA                           & U2F   & UMF   & UMCFuse                                \\ \hline
{{mAP@0.5}} & 0.794 & \cellcolor[HTML]{BDD7EE}0.817 & 0.827                         & 0.808 & 0.798 & \cellcolor[HTML]{9AFF99}\textbf{0.836} \\
mIoU                                 & 71.47 & 72.03                         & \cellcolor[HTML]{BDD7EE}75.24 & 71.36 & 70.99 & \cellcolor[HTML]{9AFF99}\textbf{75.46} \\ \hline
\end{tabular}
\label{tab_xiayou}
\end{table}
\begin{figure}[t]
  \centering
   \includegraphics[width=1.0\linewidth]{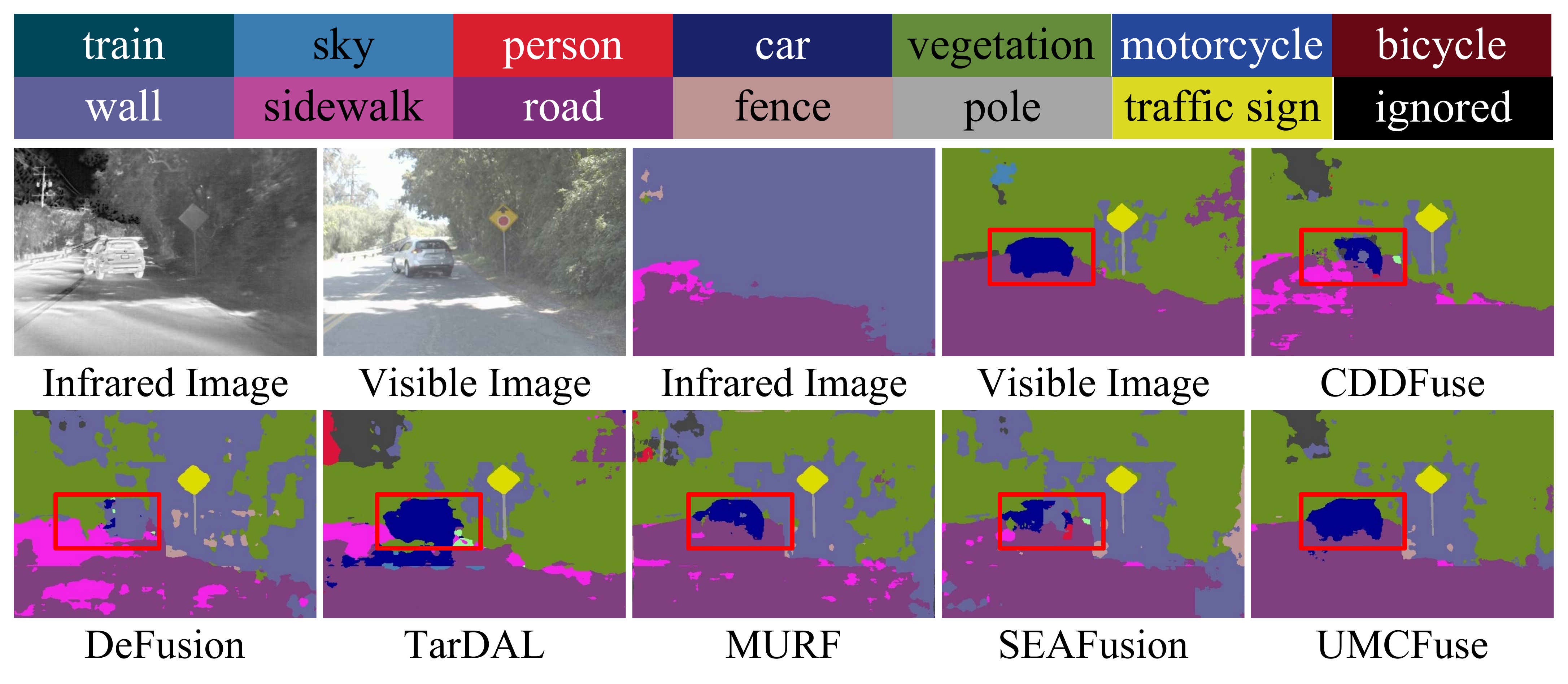}
   \caption{Visual comparison on semantic segmentation.}
   \label{fig6}
\end{figure}
In summary, the quantitative results demonstrate that the proposed algorithm excels in terms of image information entropy ($Q_{MI}$, $Q_{NCIE}$), image gradient ($Q_{G}$), structural similarity ($SSIM$), and human visual perception metrics ($Q_{CV}$, $VIF$). These metrics not only reflect better fusion quality but also indicate a greater richness in scene and semantic information, an essential factor for downstream tasks such as object detection, salient object detection, and semantic segmentation. This reflects the fact that high quality fused images help extract more accurate features, ultimately enhancing the performance of subsequent applications.

\begin{figure*}[t]
  \centering
   \includegraphics[width=1\linewidth]{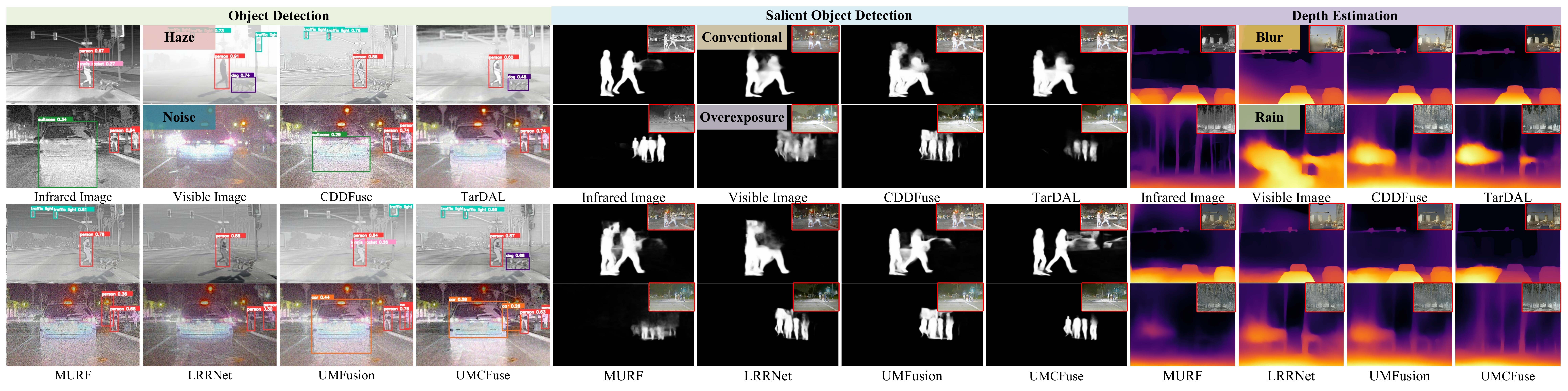}
   \caption{Visualisation results of different algorithms on three downstream tasks, object detection, salient object detection, and depth estimation.}
   \label{fig5}
\end{figure*}

\begin{table}[t]
\caption{Quantitative comparison results for different KL divergence thresholds. \textbf{Bold} is the best.}
\begin{tabular}{c|cccccc}
\hline
Para. & $Q_{MI}$↑    & $Q_{NCIE}$↑                                                                      & $Q_{G}$↑                                                                         & $Q_{abf}$↑                                                         & $Q_{CV}$↓  & $VIF$↑                                     \\ \hline
$0.01$  & \multicolumn{1}{c|}{0.3382}                                  & \multicolumn{1}{c|}{0.8063}                                  & \multicolumn{1}{c|}{0.2581}                                  & \multicolumn{1}{c|}{0.3407}                                  & \multicolumn{1}{c|}{\textbf{883.0705}} & 0.2521                                  \\
$0.05$  & \multicolumn{1}{c|}{\textbf{0.3680}} & \multicolumn{1}{c|}{\textbf{0.8067}} & \multicolumn{1}{c|}{\textbf{0.2625}} & \multicolumn{1}{c|}{\textbf{0.3425}} & \multicolumn{1}{c|}{883.5034}                                  & \textbf{0.2582} \\
$0.10$   & \multicolumn{1}{c|}{0.3535}                                  & \multicolumn{1}{c|}{0.8065}                                  & \multicolumn{1}{c|}{0.2612}                                  & \multicolumn{1}{c|}{0.3416}                                  & \multicolumn{1}{c|}{885.4119}                                  & 0.2572                                  \\ \hline
\end{tabular}
\label{tabKL}
\end{table}
\begin{table}[t]
\caption{Quantitative comparison for different smoothing factor $r$. \textbf{Bold} is the best.}
\begin{tabular}{c|cccccc}
\hline
                        & $Q_{MI}$↑    & $Q_{NCIE}$↑                                                                      & $Q_{G}$↑                                                                         & $Q_{abf}$↑                                                         & $Q_{CV}$↓  & $VIF$↑                                     \\ \cline{2-7} 
\multirow{-2}{*}{Para.} & \multicolumn{6}{c}{Overexposure}                                                                                                                                                                                                                                                                                                                                      \\ \hline
r=1                     & \multicolumn{1}{c|}{\textbf{0.3543}} & \multicolumn{1}{c|}{\textbf{0.8064}} & \multicolumn{1}{c|}{\textbf{0.4774}} & \multicolumn{1}{c|}{0.4478}                                  & \multicolumn{1}{c|}{\textbf{754.1789}}  & \textbf{0.2388} \\
r=2                   & \multicolumn{1}{c|}{0.3530}                                  & \multicolumn{1}{c|}{\textbf{0.8064}} & \multicolumn{1}{c|}{0.4763}                                  & \multicolumn{1}{c|}{0.4505}                                  & \multicolumn{1}{c|}{761.0015}                                   & 0.2383                                  \\
r=3                     & \multicolumn{1}{c|}{0.3506}                                  & \multicolumn{1}{c|}{0.8063}                                  & \multicolumn{1}{c|}{0.4735}                                  & \multicolumn{1}{c|}{\textbf{0.4518}} & \multicolumn{1}{c|}{768.4901}                                   & 0.2373                                  \\
r=4                     & \multicolumn{1}{c|}{0.3493}                                  & \multicolumn{1}{c|}{0.8063}                                  & \multicolumn{1}{c|}{0.4742}                                  & \multicolumn{1}{c|}{0.4506}                                  & \multicolumn{1}{c|}{773.1851}                                   & 0.2377                                  \\ \hline
\multicolumn{1}{l|}{}   & \multicolumn{6}{c}{Noise}                                                                                                                                                                                                                                                                                                                                             \\ \hline
r=1                     & \multicolumn{1}{c|}{\textbf{0.2977}} & \multicolumn{1}{c|}{\textbf{0.8058}} & \multicolumn{1}{c|}{\textbf{0.1904}} & \multicolumn{1}{c|}{\textbf{0.2671}} & \multicolumn{1}{c|}{\textbf{1093.5527}} & \textbf{0.2033} \\
r=2                     & \multicolumn{1}{c|}{0.2621}                                  & \multicolumn{1}{c|}{0.8053}                                  & \multicolumn{1}{c|}{0.1576}                                  & \multicolumn{1}{c|}{0.2286}                                  & \multicolumn{1}{c|}{1235.0823}                                  & 0.1778                                  \\
r=3                     & \multicolumn{1}{c|}{0.2407}                                  & \multicolumn{1}{c|}{0.8050}                                  & \multicolumn{1}{c|}{0.1391}                                  & \multicolumn{1}{c|}{0.2072}                                  & \multicolumn{1}{c|}{1335.7768}                                  & 0.1646                                  \\
r=4                     & \multicolumn{1}{c|}{0.2402}                                  & \multicolumn{1}{c|}{0.8049}                                  & \multicolumn{1}{c|}{0.1337}                                  & \multicolumn{1}{c|}{0.1953}                                  & \multicolumn{1}{c|}{1400.1665}                                  & 0.1589                                  \\ \hline
\end{tabular}
\label{tabr}
\end{table}

\subsection{Downstream Tasks Experiments}
We evaluate the different methods in seven scenes, and four downstream tasks cover semantic segmentation \cite{r58}, object detection \cite{r9}, salient object detection \cite{r59}, and depth estimation \cite{r60}. As shown in \cref{fig6} and \cref{fig5}, UMCFuse has the best performance in all experiments, proving that it can facilitate downstream tasks under different interferences, integrating useful information from different modal images while being adaptive to interfering pixels.
Regarding object detection, UMCFuse exhibits the highest number of detected objects and demonstrates a high confidence level. This indicates the proposed method's significant impact on object detection and its ability to enhance scene interpretation by integrating valuable information from different modalities.
In the task of salient object detection, it can be seen that, except for UMCFuse, none of the algorithms can effectively and accurately detect the location information of the pedestrians, and all of them show different degrees of artifacts and incomplete detection.
In addition, in the depth estimation experiments, the comparison methods result in the failure of downstream algorithms to accurately measure scene depth due to the retention of smoke and ambiguity information.

Furthermore, we present the results of a quantitative comparison of all methods on the M3FD dataset \cite{r3} for the object detection task \cite{r10} and on the MSRS dataset \cite{r94} for the semantic segmentation task \cite{r93}. As shown in \cref{tab_xiayou}, the proposed algorithm achieves the highest accuracy in detection and segmentation, further validating its effectiveness in downstream tasks. These results demonstrate the potential of the proposed algorithm to enhance machine understanding and scene analysis.

\begin{figure}[t]
  \centering
   \includegraphics[width=1.0\linewidth]{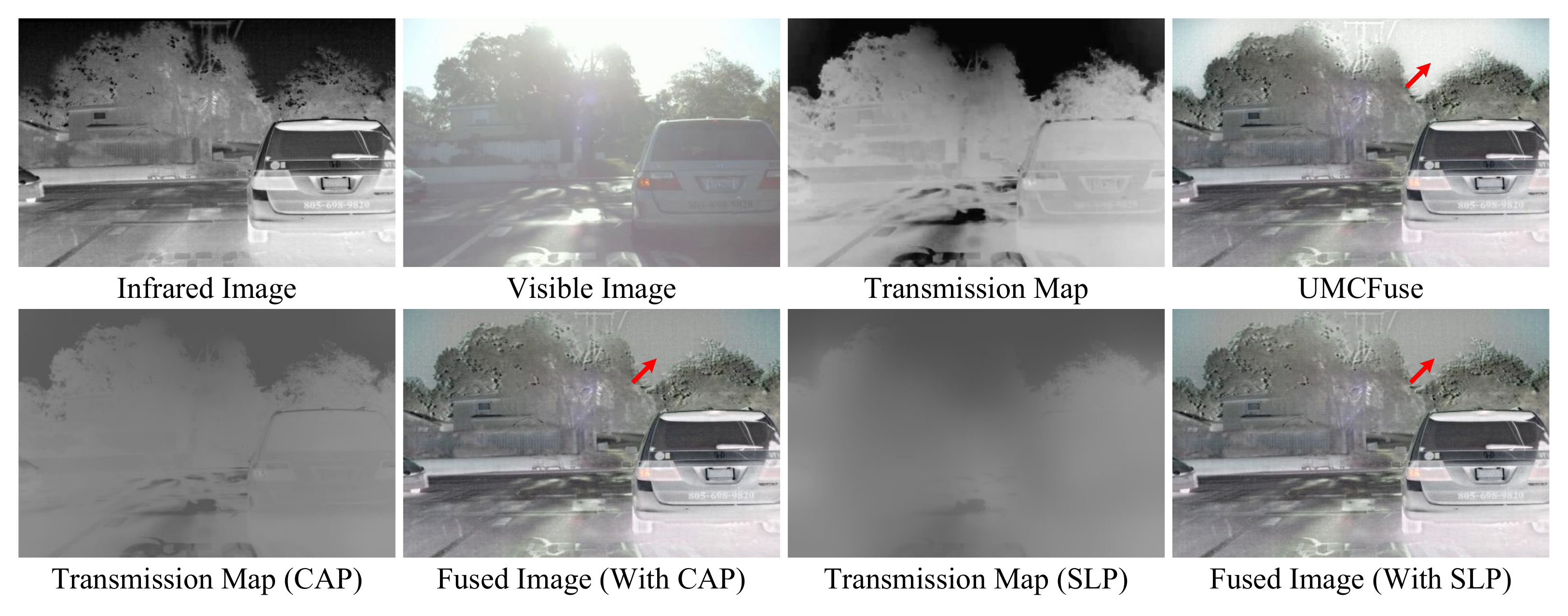}
   \caption{Fusion results obtained by different transmission maps.}
   \label{map}
\end{figure}

\subsection{Parametric Analysis}
In order to assess the impact of different hyperparameters on the fusion performance of the proposed algorithm, we analyzed the KL divergence thresholds and the smoothing factor $r$. \cref{tabKL} presents the quantitative comparison results for various KL divergence thresholds, using 50 randomly selected images from the RoadScene dataset. We observe that the proposed algorithm achieves the best fusion performance when the threshold is set to $0.05$.

The smoothing coefficient $r$ also significantly affects the proposed algorithm. This coefficient is primarily used for the high and low-frequency decomposition of images. We randomly selected 50 images from two different scenes and analyzed the results for different values of $r$. In the overexposure scene, a larger $r$ results in excessive smoothing of the contrast layer and the low-frequency components of the infrared image. This leads to an overemphasis on the overexposed regions and thermal radiation information, which is detrimental to the fusion process, as it fails to suppress regions of excessively high contrast. Similarly, in noisy scenes, a higher value of $r$ causes more detail and noise to be separated from the low-frequency components, with noise suppression becoming overly reliant on high-frequency processing. This can result in significant detail loss in the fusion outcome or leave residual noise in the final image. Based on the quantitative comparison results in \cref{tabr}, we conclude that the proposed algorithm performs best when $r=1$.

\begin{table*}[t]
\caption{Quantitative comparison of all methods on SPECT-MRI fusion task. Green is the best and blue is the second.}
\centering
\begin{tabular}{c|cccccccc}
\hline
Methods   & $Q_{MI}$↑                                                          & $Q_{NCIE}$↑                                                        & $Q_{G}$↑                                                           & $Q_{abf}$↑                                                         & $Q_{CV}$↓                                                            & $VIF$↑                                                          & $EN$↑                                                           & $SSIM$↑                                                    \\ \hline
CDDFuse \cite{r2}    & \multicolumn{1}{c|}{\cellcolor[HTML]{DDEBF7}0.7142}          & \multicolumn{1}{c|}{0.8083}                                  & \cellcolor[HTML]{9AFF99}\textbf{0.6917}             & \multicolumn{1}{c|}{0.6320}                                  & \multicolumn{1}{c|}{231.1441}                                  & \multicolumn{1}{c|}{0.3211}                                  & \multicolumn{1}{c|}{\cellcolor[HTML]{DDEBF7}5.0793}          & \cellcolor[HTML]{DDEBF7}0.6154          \\

SwinFusion \cite{r19} & \multicolumn{1}{c|}{0.6565}                                  & \multicolumn{1}{c|}{0.8078}                                  & \multicolumn{1}{c|}{0.6128}                         & \multicolumn{1}{c|}{0.6409}                                  & \multicolumn{1}{c|}{\cellcolor[HTML]{DDEBF7}138.7817}          & \multicolumn{1}{c|}{0.3222}                                  & \multicolumn{1}{c|}{\cellcolor[HTML]{9AFF99}\textbf{5.3393}} & 0.2520                                  \\

PRRGAN \cite{r88}     & \multicolumn{1}{c|}{\cellcolor[HTML]{9AFF99}\textbf{0.7789}} & \multicolumn{1}{c|}{\cellcolor[HTML]{DDEBF7}0.8095}          & \multicolumn{1}{c|}{0.6899}                         & \multicolumn{1}{c|}{\cellcolor[HTML]{DDEBF7}0.6493}          & \multicolumn{1}{c|}{\cellcolor[HTML]{9AFF99}{\textbf{105.6865}}} & \multicolumn{1}{c|}{\cellcolor[HTML]{DDEBF7}{0.3529}} & \multicolumn{1}{c|}{4.8707}                                  & 0.6076                                  \\

MDHU \cite{r89}       & \multicolumn{1}{c|}{0.7049}                                  & \multicolumn{1}{c|}{0.8080}                                  & \multicolumn{1}{c|}{0.6767}                         & \multicolumn{1}{c|}{0.5787}                                  & \multicolumn{1}{c|}{344.2878}                                  & \multicolumn{1}{c|}{0.2584}                                  & \multicolumn{1}{c|}{5.0586}                                  & 0.6099                                  \\ \hline

UMCFuse    & \multicolumn{1}{c|}{0.6895}                                  & \multicolumn{1}{c|}{\cellcolor[HTML]{9AFF99}\textbf{0.8097}} & \multicolumn{1}{c|}{\cellcolor[HTML]{DDEBF7}0.6905} & \multicolumn{1}{c|}{\cellcolor[HTML]{9AFF99}\textbf{0.6526}} & \multicolumn{1}{c|}{161.0538}                                  & \multicolumn{1}{c|}{\cellcolor[HTML]{9AFF99}\textbf{0.3561}}          & \multicolumn{1}{c|}{4.9033}                                  & \cellcolor[HTML]{9AFF99}\textbf{0.6173} \\ \hline
\end{tabular}
\label{tab3}
\end{table*}

\begin{figure}[t]
  \centering
   \includegraphics[width=1.0\linewidth]{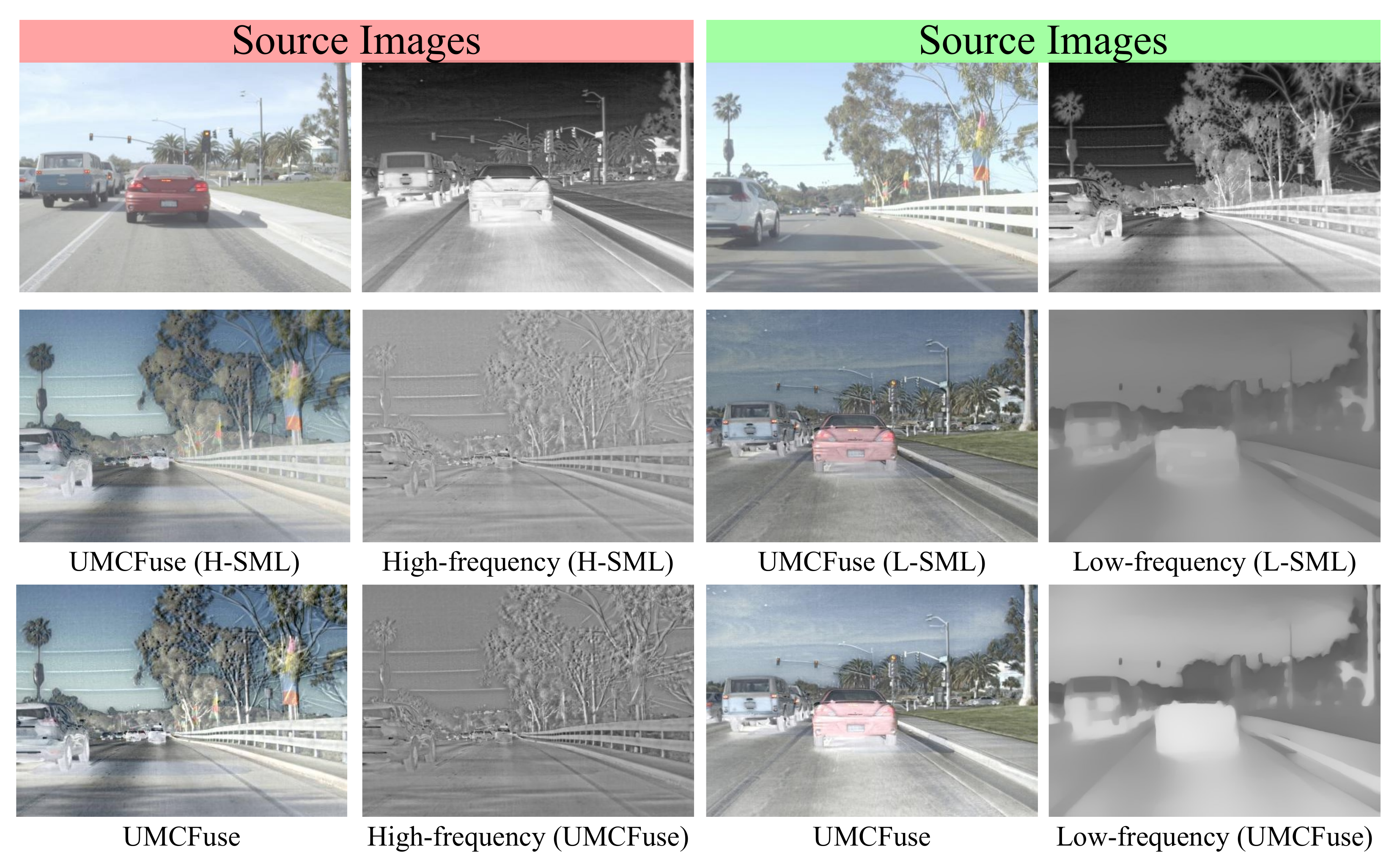}
   \caption{Fusion results obtained by different fusion rules.}
   \label{rule}
\end{figure}

\subsection{Ablation Experiments}
\noindent \textbf{Transmission Map Estimation:}
The transmission map is crucial for guiding image decomposition and significantly impacts the light transmission perception of the proposed algorithm. It determines the information contained in the structure and contrast layers. To validate the effectiveness of the proposed transmission map estimation strategy, we compare it with other priori methods, specifically the Saturation Line Prior (SLP) \cite{r91} and the Color Attenuation Prior (CAP) \cite{r92}. We perform image decomposition by replacing the transmission map estimated by our algorithm with those obtained from these alternative methods. As shown in \cref{map}, we observe that the transmission maps obtained using the SLP and CAP methods are blurred and fail to capture the critical structural information in the images. Furthermore, fusion results based on these transmission maps exhibit a loss of contrast, such as the inability to preserve the energy information in the sky.

\noindent \textbf{Fusion Rule:}
To evaluate the influence of the high-frequency and low-frequency fusion rules on the performance of the proposed algorithm, we perform a series of ablation experiments. Initially, we replace the MPC operator in the high-frequency fusion rule with the Sum-modified-Laplacian measure (SML), which results in the method we refer to as UMCFuse (H-SML). Subsequently, we substitute the feature map derived from SML for the feature measurement map $V_{\theta}^{\gamma}$ in the low-frequency fusion rule, resulting in the method UMCFuse (L-SML). The fusion results, along with the high-frequency and low-frequency images for both ablation experiments and the proposed algorithm, are illustrated in \cref{rule}.  As indicated in the figure, the high-frequency component fused with SML displays a loss of detail, leading to a decrease in contrast in the fusion output. In contrast, the low-frequency component fused using SML shows a decrease in image energy, which results in less prominent target information in the final results.

\subsection{Detail Loss Judgment Effectiveness Analysis}
\begin{figure*}[t]
  \centering
   \includegraphics[width=1\linewidth]{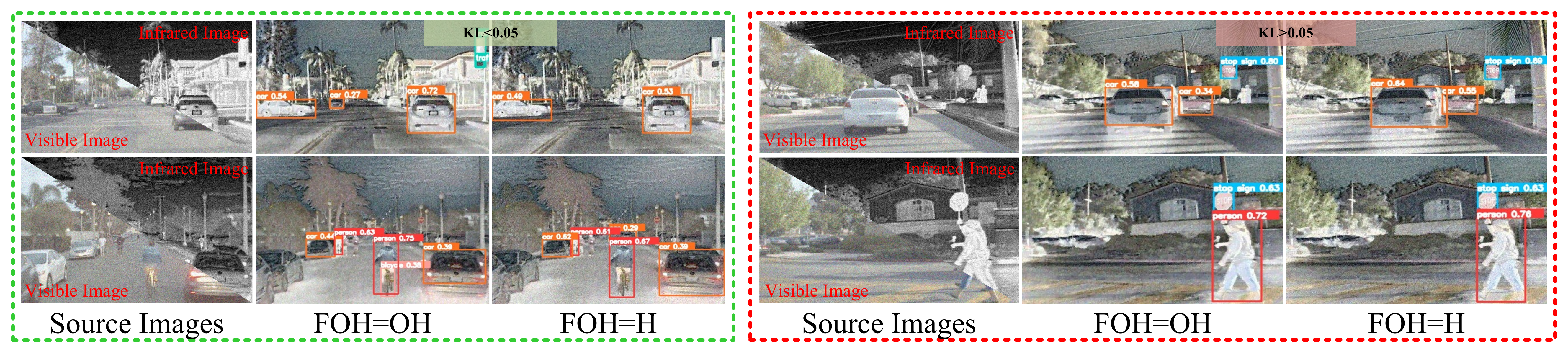}
   \caption{Visualization experiments for effectiveness analysis of detail loss judgment.}
   \label{fig8}
\end{figure*}

In high-frequency component fusion, we question whether denoising inevitably leads to detail loss. When the image experiences minor noise, is denoising necessary? In our study, we set a judgment condition based on KL divergence. As shown in \cref{fig8}, when $KL < 0.0$5, the proposed algorithm judges it as the degree of detail loss after denoising is less than the effect of noise on the detail distribution, and it can also be found in the object detection experiments that the information of the significant target will be more prominent when $F = OH$. When $KL > 0.05$, discarding the denoising process so that $F = H$ will get higher quality results.

\subsection{Extended Experiments on Medical Image Fusion Task}
In addition to validating the fusion performance of the proposed algorithm on infrared and visible modalities, we have extended our experiments to the medical image fusion task. For this purpose, we randomly selected 50 sets of SPECT-MRI images from the Harvard Medical School database. We compared the proposed method against two generalized multi-modal image fusion methods (CDDFuse\cite{r2} and SwinFusion\cite{r19}) as well as two medical image fusion methods (PRRGAN\cite{r88} and MDHU\cite{r89}). The quantitative results, presented in \cref{tab3}, demonstrate that the proposed algorithm ranks in the top two across all five metrics. Moreover, although medical images do not conform to the ASM typically used in imaging, the proposed method is still able to extract significant features and achieve excellent fusion results, highlighting its robust adaptability to cross-domain datasets.

\begin{figure}[h]
  \centering
   \includegraphics[width=1.0\linewidth]{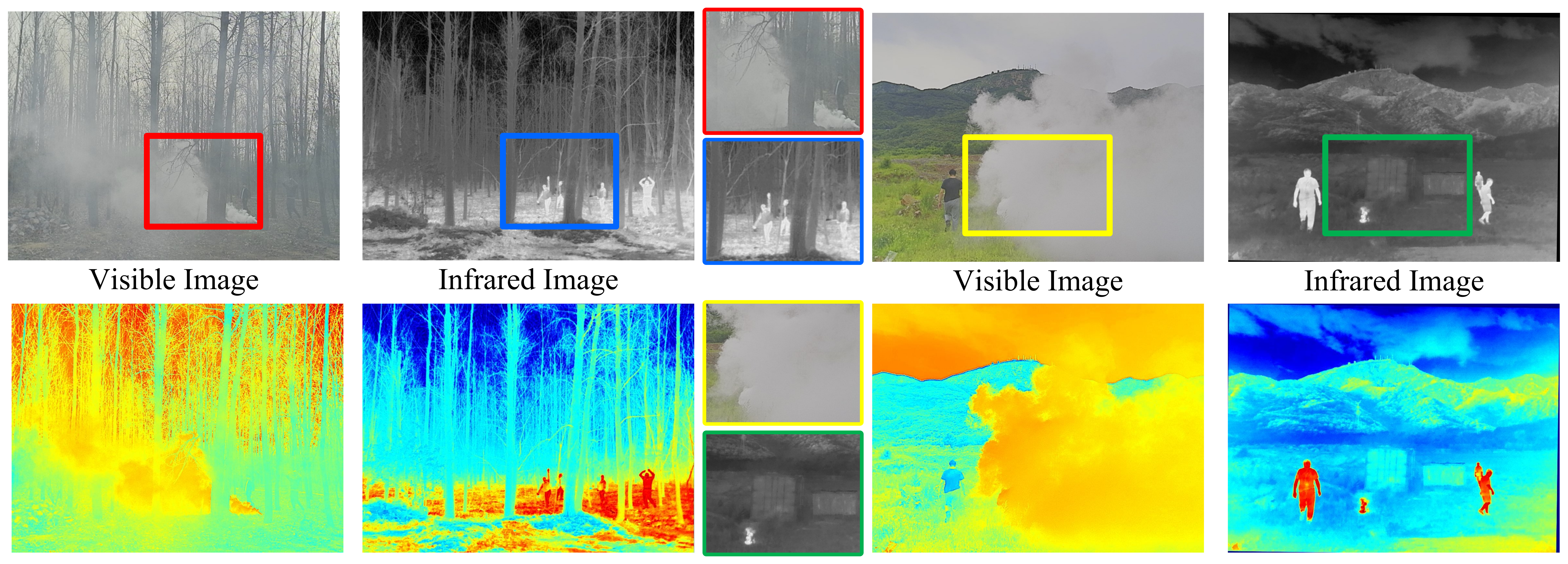}
   \caption{Visible and infrared images and pseudo-colour maps in haze scenes.}
   \label{fig9}
\end{figure}

\subsection{Computational Efficiency Analysis}

In this section, we provide a comparative analysis of the computational efficiency of different algorithms in terms of inference time (TIME) and Floating Point Operations per Second (FLOPs). All the experiments in this section are performed on images with a resolution of $480\times640$. As shown in the \cref{time}, it can be noticed that the DL-based methods are implemented on a GPU (RTX 3090Ti), and these methods consistently outperform others in terms of inference time due to the parallel computing and accelerated processing capabilities of GPUs. For example, TarDAL and SeaFusion methods are both able to meet real-time processing requirements with GPU acceleration. In contrast, two methods, PFF and UMCFuse, are implemented through the Matlab architecture, which performs the computation through CPU (Core i7-13700F), which itself is weaker than the GPU. Therefore, the comparison of inference time is not fair to the traditional methods based on the CPU. In addition, the proposed algorithm uses techniques that cover multi-component decomposition and multiscale computation, which inevitably suffer from common inefficiencies. 
Additionally, since none of the compared methods can be directly applied to complex scenes, they require image restoration as a pre-processing step to handle interference caused by adverse weather conditions or exposure anomalies. Some classical image restoration algorithms, such as Restormer, introduce additional overhead, adding $379.163\,\text{ms}$ of inference time and $87.7\,\text{GFLOPs}$ of extra computational cost when integrated with image fusion algorithms.

To offer a more objective and in-depth evaluation of the computational efficiency of the different algorithms, FLOPs have been carried out. From the \cref{time}, it can be found that although UMCFuse does not significantly outperform other methods in inference time, it possesses the smallest GFLOPs, indicating that it requires fewer computational resources and less memory for image processing. This makes UMCFuse more suitable for applications in environments with limited computational resources, where minimizing overhead is crucial. Overall, the proposed algorithm can be transformed into a more efficient implementation by optimizing the code, such as GPU acceleration or implementing the code in \texttt{C++}, which can satisfy more real-world problems.

\begin{table}[t]
\caption{Comparative Analysis of Computational Efficiency Across Different Methods: Inference Time (ms) and FLOPs (G).}
\centering
\begin{tabular}{c|c|c|c|c}
\hline
Methods   & Hardware                         & Framework                    & TIME                              & FLOPs                              \\ \hline
CDDFuse\cite{r2}   &                                  &                              & 603                                 & 293                                \\
DeFusion\cite{r39}  &                                  &                              & 429                                 & 91.52                                 \\
ReCoNet\cite{r41}   &                                  &                              & \cellcolor[HTML]{FFCCC9}34.8          & \cellcolor[HTML]{DDEBF7}7.12          \\
TarDAL\cite{r3}    &                                  &                              & \cellcolor[HTML]{9AFF99}\textbf{10.9} & 51.34                                 \\
LRRNet\cite{r16}    &                                  &                              & 91.9                                  & \cellcolor[HTML]{FFCCC9}7.98          \\
SeAFusion\cite{r4} &                                  &                              & \cellcolor[HTML]{DDEBF7}17.6          & 28.73                                 \\
UMFusion\cite{r20}  & \multirow{-7}{*}{RTX3090Ti}     & \multirow{-7}{*}{Pytorch}    & 82.6                                 & 193                                \\ \hline
DIVFusion\cite{r17} &                                  &                              & 1018                                & 836                                \\
MURF\cite{r42}      &                                  &                              & 548                                 & 126                                \\
U2Fusion\cite{r34}  & \multirow{-3}{*}{RTX3090Ti}     & \multirow{-3}{*}{Tensorflow} & 3152                                   & 1383                               \\ \hline
PFF\cite{r22}       &                                  &                              & 4412                                & 11.41                                 \\
UMCFuse   & \multirow{-2}{*}{Core i7-13700F} & \multirow{-2}{*}{Matlab}     & 1353                                   & \cellcolor[HTML]{9AFF99}\textbf{5.04} \\ \hline
\end{tabular}
\label{time}
\end{table}

\subsection{Feasibility Analysis of the Proposed Method in the Real-World}
There is no IVIF dataset currently that comprehensively covers real-world adverse weather scenes (e.g., heavy rain, haze, and snow) and challenging conditions. In adverse weather scenes, the scattering induced by raindrops, haze, and snow, along with the heat they dissipate while traversing through the air, collectively diminishes the effective signal received by the infrared sensor, resulting in contrast degradation and blurring. It is worth noting that the specific features of severe weather are not replicated in the infrared images \cite{r70}, they only cause degradation to the quality of the infrared imaging (as shown in \cref{fig9}). Hence, we focused on eliminating severe weather-specific features from visible images. For infrared images, which lack weather features, our focus has been on capturing their valuable information. Overall, they make the proposed algorithm applicable to complex scenes in the real-world.

\subsection{Limitations and Future Work}
\noindent \textbf{Limitations:} For the first time, we introduce a unified framework for IVIF in complex scenes. Although the proposed method may not consistently exceed the performance of ``restoration + fusion'' combinations in certain specific tasks, it offers significant advantages by providing a consistent solution to address multiple image fusion challenges within a single framework. The reasons are as follows: (1) The proposed algorithm is the first to incorporate light transmission information to guide the fusion process in complex scenes. Comprehensive experiments have validated the applicability of this strategy across various interference factors, thereby offering a novel perspective for developing more comprehensive models in the future. (2) To the best of our knowledge, no existing method can simultaneously address image fusion in complex scenes with uniform weights for multiple types of interference. However, the ``restoration + fusion'' strategy is still suboptimal, since manually choosing a suitable restoration method per scene is inefficient and often infeasible.  

\noindent \textbf{Future Work:} In the future, we will extend our algorithm to support both real-time and misaligned image fusion in complex scenes. Our unified framework provides a comprehensive view of the IVIF problem in challenging environments, which can enable subsequent research to focus on how to simplify the computation and quickly isolate image degradation factors. For misaligned image fusion in complex scenes, this degradation separation method can help reduce the identification of erroneous feature points, thereby improving the alignment accuracy of multi-modality images.

\section{Conclusion}
In this study, we propose an IVIF method for complex scenes, which can eliminate the interference while obtaining outstanding fusion performance. Specifically, we propose a decomposition model based on transmittance estimation, and from the perspective of atmospheric light scattering, to realize interfering pixels classification. Given that high-frequency components are highly sensitive to environmental changes and noise, we propose an adaptive denoising fusion strategy that suppresses noise while preserving important details. For low-frequency components, the algorithm effectively detects energy information in high-contrast regions through multi-scale, multi-directional feature extraction. Extensive experiments show that the proposed algorithm produces superior fusion results even under single or multiple interferences.


\bibliographystyle{IEEEtran}
\bibliography{main}

\vspace{-33pt}

\begin{IEEEbiography}[{\includegraphics[width=1in,height=1.25in,clip,keepaspectratio]{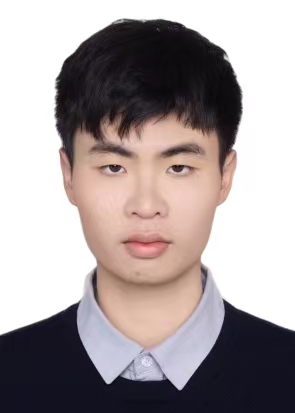}}]{Xilai Li}
received his Bachelor of Engineering degree from the School of Physics and Optoelectronic Engineering at Foshan University. His current research interests include image processing and machine learning.
\end{IEEEbiography}
\begin{IEEEbiography}[{\includegraphics[width=1in,height=1.25in,clip,keepaspectratio]{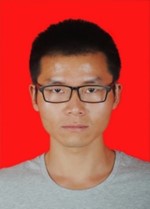}}]{Xiaosong Li}
received his Ph.D. in measurement technology and instruments from Beihang University in 2021. He is currently an associate professor with the School of Physics and Optoelectronic Engineering, Foshan University, Foshan, China. His research interests include image processing, pattern recognition, and optical metrology.
\end{IEEEbiography}
\begin{IEEEbiography}[{\includegraphics[width=1in,height=1.25in,clip,keepaspectratio]{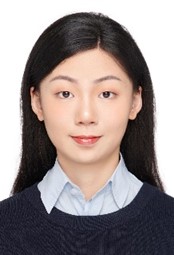}}]{Tianshu Tan}
received her Bachelor of Engineering degree in Computer Engineering and Artificial Intelligence from the Hong Kong University of Science and Technology. Her research interest include Brain Computer Interface, Machine Learning and Image Processing.
\end{IEEEbiography}
\begin{IEEEbiography}[{\includegraphics[width=1in,height=1.25in,clip,keepaspectratio]{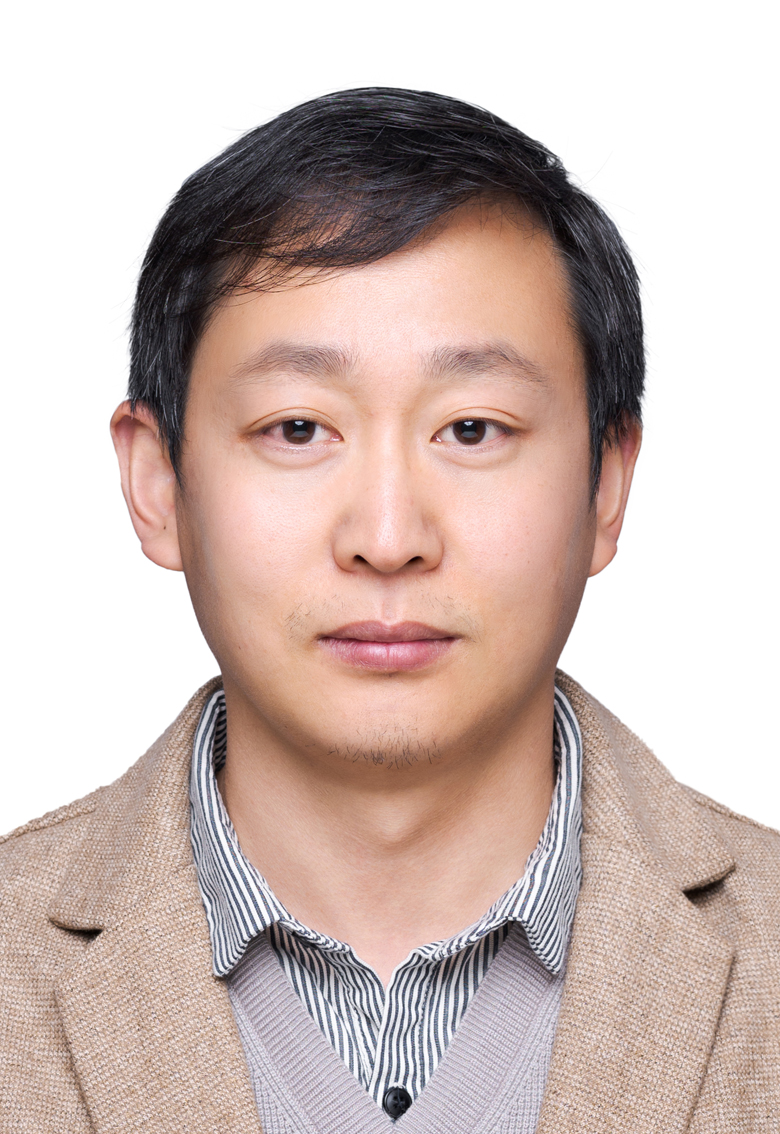}}]{Huafeng Li}
received the M.S. degree in applied mathematics major from Chongqing University in 2009 and obtained his Ph.D. degree in control theory and control engineering major from Chongqing University in 2012.  He is currently a professor at the School of Information Engineering and Automation, Kunming University of Science and Technology, China. His research interests include image processing, computer vision, and information fusion.
\end{IEEEbiography}
\begin{IEEEbiography}[{\includegraphics[width=1in,height=1.25in,clip,keepaspectratio]{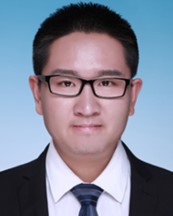}}]{Tao Ye}
completed Ph.D. in measurement technology and instruments from Beihang University, Beijing, in 2016. From March 2016 to March 2019, he was an Engineer with the Beijing Institute of Remote Sensing and Equipment, Beijing. He is currently a Senior Engineer with the School of Mechanical Electronic and Information Engineering, China University of Mining and Technology. His current research interests include deep learning and traffic detection.
\end{IEEEbiography}

\vspace{-33pt}

\end{document}